\documentclass[12pt]{elsarticle}

\usepackage[english]{babel}
\usepackage{graphicx}
\usepackage{dcolumn}
\usepackage{bm}
\usepackage{amsmath,amsfonts,amssymb}
\usepackage{hyperref}
\usepackage{mathtools}
\usepackage{multirow}
\usepackage[table]{xcolor}
\usepackage{natbib}
\usepackage{colortbl}
\usepackage{regexpatch}
\makeatletter
\regexpatchcmd\ps@pprintTitle
  {\cE\}\ \cE\}}{\cE\}\cE\}}
  {}{\FailedToPatch}
\makeatother
\newcommand{\vv}[1]{\textbf{#1}}
\newcommand{\bb}[1]{\boldsymbol{#1}}
\newcommand{\di}[1]{\textrm{diag}(#1)}
\newcommand{\overbar}[1]{\mkern 1.5mu\overline{\mkern-1.5mu#1\mkern-1.5mu}\mkern 1.5mu}
\makeatletter
\def\ps@pprintTitle{%
  \let\@oddhead\@empty
  \let\@evenhead\@empty
  \def\@oddfoot{\reset@font\hfil\thepage\hfil}
  \let\@evenfoot\@oddfoot
}
\makeatother

\begin{document}

\title{Stabilizing Machine Learning Prediction of Dynamics: Noise and Noise-inspired Regularization}

\author[phys]{Alexander Wikner\corref{cor1}}
\ead{awikner1@umd.edu}
\author[hill]{Joseph Harvey\texorpdfstring{\fnref{fn1}}}
\author[phys]{Michelle Girvan}
\author[math]{Brian R. Hunt}
\author[prl]{Andrew Pomerance}
\author[phys]{Thomas Antonsen}
\author[phys,elec]{Edward Ott}
\cortext[cor1]{Corresponding author}
\fntext[fn1]{Joseph Harvey is currently employed at Aunalytics in South Bend, IN 46601.}
\affiliation[phys]{organization = {Department of Physics, University of Maryland},
addressline={4150 Campus Dr},
postcode={20742},
city={College Park},
country={United States}}
\affiliation[math]{organization = {Department of Mathematics, University of Maryland},
addressline={4176 Campus Dr},
postcode={20742},
city={College Park},
country={United States}}
\affiliation[elec]{organization = {Department of Electrical and Computer Engineering, University of Maryland},
addressline={8223 Paint Branch Dr},
postcode={20742},
city={College Park},
country={United States}}
\affiliation[hill]{organization = {Hillsdale College},
addressline={33 E College St},
postcode={49242},
city={Hillsdale},
country={United States}}
\affiliation[prl]{organization = {Potomac Research LLC},
addressline={801 N Pitt St},
postcode = {22341},
city={Alexandria},
country={United States}}

\date{\today}

\begin{abstract}
Recent work has shown that machine learning (ML) models can be trained to accurately forecast the dynamics of unknown chaotic dynamical systems. Short-term predictions of the state evolution and long-term predictions of the statistical patterns of the dynamics (``climate'') can be produced by employing a feedback loop, whereby the model is trained to predict forward one time step, then the model output is used as input for multiple time steps. In the absence of mitigating techniques, however, this technique can result in artificially rapid error growth. In this article, we systematically examine the technique of adding noise to the ML model input during training to promote stability and improve prediction accuracy. Furthermore, we introduce Linearized Multi-Noise Training (LMNT), a regularization technique that \textit{deterministically} approximates the effect of many small, independent noise realizations added to the model input during training. Our case study uses reservoir computing, a machine learning method using recurrent neural networks, to predict the spatiotemporal chaotic Kuramoto-Sivashinsky equation. We find that reservoir computers trained with noise or with LMNT produce climate predictions that appear to be indefinitely stable and have a climate very similar to the true system, while reservoir computers trained without regularization are unstable. Compared with other regularization techniques that yield stability in some cases, we find that both short-term and climate predictions from reservoir computers trained with noise or with LMNT are substantially more accurate. Finally, we show that the deterministic aspect of our LMNT regularization facilitates fast hyperparameter tuning when compared to training with noise.
\end{abstract}


\maketitle
\section{Introduction}\label{sec:intro}

Learning dynamics solely from state time-series measurements of otherwise unknown complex dynamical systems is a challenging problem for which, in recent years, machine learning (ML) has been shown to be a promising solution. For example, ML models trained on time series measurements have been applied to obtain accurate predictions of terrestrial weather~\cite{rasp_weatherbench_2020,arcomano_machine_2020, rasp_data-driven_2021, pathak_fourcastnet_2022}. Chaotic dynamics is of particular interest due to its common occurrence in complex systems. Time series measurements from such systems have been accurately predicted using ML (e.g., in Refs.~\cite{pathak_model-free_2018,vlachas_data-driven_2018, vlachas_backpropagation_2020,balakrishnan_deep_2020}), although, due to the exponentially sensitive dependence of chaotic orbits on perturbations, the time duration for which specific measurements can be predicted is necessarily limited (e.g., as in weather forecasting). Nonetheless, ML models can also produce arbitrarily long-term predictions that approximate the correct ``climate''~\cite{rasp_deep_2018, gentine_could_2018}, by which we mean a statistical description of the long-term system behavior. However, as with numerical methods for solving differential equations, ML models sometimes generate artificial instabilities that lead to an inaccurate climate. We call this situation a ``climate instability''; such an instability might or might not degrade the accuracy of a short term forecast.
\subsection{Machine learning prediction of dynamics and the issue of climate stability}{\label{sec:ml_prediction}}
\begin{figure}[!ht]
    \centering
    \includegraphics[width = \textwidth]{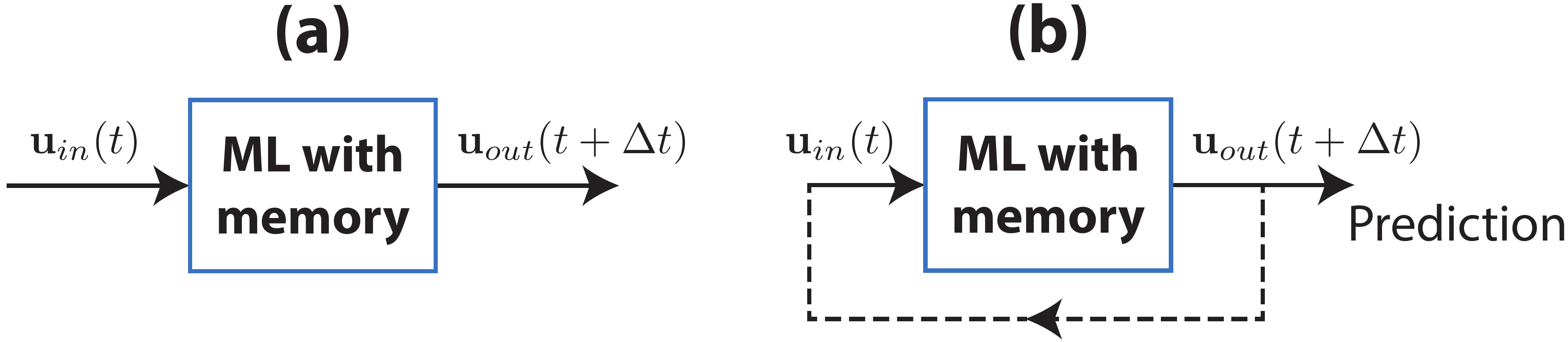}
    \caption{Machine Learning Model for Predicting Dynamics. The above diagrams show an ML model for prediction in the (a) ``open-loop'' configuration (for training) and (b) ``closed-loop'' configuration (for prediction). In panel (b), the dashed line indicates that the output is fed back into the model as the next input.}
    \label{fig:basicML}
\end{figure}

In this article, we consider the commonly employed ML scheme used for predicting dynamical system states from time-series measurements $\{\vv{u}(t)\}$, the general structure of which is shown in Fig.~\ref{fig:basicML}. We note that we are concerned here with ML models that contain some form of a ``memory'' of previous inputs, so that $\vv{u}_{out}(t+\Delta t)$ depends not only on $\vv{u}_{in}(t)$, but also on $\vv{u}_{in}(t-\Delta t)$, $\vv{u}_{in}(t-2\Delta t)$, etc. This memory is a necessary component when one does not have access to full system measurements, and it often improves model performance even when full system state measurements are available. The goal during ML model training in this scheme (Fig.~\ref{fig:basicML}(a)) is to adjust the model weights such that when the model input is $\vv{u}_{in}(t) = \vv{u}(t)$, the output $\vv{u}_{out}(t+\Delta t)$ closely matches the measurements $\vv{u}(t+\Delta t)$. We refer to Fig.~\ref{fig:basicML}(a) as the ``open-loop'' configuration. After training, we initialize prediction after time $t=T_{init}\Delta t$ by switching the model to the configuration shown in Fig.~\ref{fig:basicML}(b), which we refer to as the ``closed-loop'' configuration. In this configuration, the output $\vv{u}_{out}((T_{init}+n)\Delta t)$ is used as the next input $\vv{u}_{in}((T_{init}+n)\Delta t)$ for $n=1,2,3,\dots$, and the outputs $\{\vv{u}_{out}((T_{init}+n)\Delta t)\}$ also form the predicted time series of measurements. This closed-loop configuration is used for prediction in Refs.~\cite{rasp_weatherbench_2020, arcomano_machine_2020,rasp_data-driven_2021, pathak_fourcastnet_2022,pathak_model-free_2018,vlachas_data-driven_2018,vlachas_backpropagation_2020,li_fourier_2021}.

Once a trained ML model has been placed into the closed-loop configuration, it acts as an autonomous dynamical system. Assuming that the unknown dynamical system generating the training data $\{\vv{u}(t)\}$ is evolving on an ``attractor'' (an invariant set of the state space dynamics that ``attracts'' nearby orbits~\cite{auslander_attractors_1964}), the closed-loop ML model plausibly has an invariant set that approximates this attractor~\cite{lu_attractor_2018}. From examples, it appears that this approximating invariant set is often indeed an attractor for the ML model, with ergodic properties nearly identical to those of the true dynamical system being measured~\cite{pathak_using_2017}. This has been demonstrated in examples where the unknown dynamical has multiple attractors~\cite{gauthier_learning_2022} and where the training data does not sample the unknown dynamical system attractor~\cite{rohm_model-free_2021}. However, even in cases where the existence of an approximating invariant set can be guaranteed, it might be that small perturbations transverse to the invariant set grow with time, so that, eventually, the ML model climate grossly differs from that of the dynamical system producing the training data $\{\vv{u}(t)\}$~\cite{lu_attractor_2018}. Climate stability requires suppressing the growth of such perturbations. In chaotic systems where accurate long-term prediction is impossible, such as the earth's atmosphere, obtaining the correct climate is often the goal of long-term predictions. In addition, climate instability can limit the duration of accurate short-term state forecasts (as shown, e.g., in Figs. 4 and 5 of Ref.~\cite{pathak_using_2017}). This can occur if the growth time of the climate instability is fast enough that it causes substantial deviation from the invariant set of the closed-loop system before the predictions break down due to the natural chaos of the orbits on the original attractor of the unknown system being predicted.

\subsection{Stabilization}\label{sec:stabilization}
With the aim of improving stability and climate prediction without sacrificing short-term prediction accuracy, one of our primary goals in this article is to systematically study the effect of added noise to the model input, as discussed in Refs.~\cite{jaeger_echo_2001,lukosevicius_reservoir_2009-1}. Roughly speaking, one can view the added noise as perturbing the input training orbit off the target invariant set of the dynamics so that, during the training, the ML model learns to respond to input from a neighborhood of the invariant set. Thus, perturbations from the invariant set are trained to be pulled back toward the invariant set, tending to make it stable (i.e., make it an attractor). We emphasize, however, that this reasoning is only heuristic, because it is based on the open-loop system (Fig.~\ref{fig:basicML}(a)), while the prediction uses the closed-loop system (Fig.~\ref{fig:basicML}(b)).

In addition to considering the utility of input noise, we also introduce a new regularization technique, Linearized Multi-Noise Training (LMNT), for training ML models with memory. While the LMNT regularization technique is based on the idea of adding noise to the input training data, LMNT is a deterministic, non-stochastic procedure. In the context of our chosen machine learning method, reservoir computing, we find that LMNT greatly simplifies the tuning of the hyperparameter associated with the regularization strength when compared to the approach of adding noise to the model input.
\subsection{Outline and main results}
Our article is structured as follows: In Sec.~\ref{sec:methods}, we describe the basic structure of a machine learning predictor that is trained to produce short-term forecasts and then used for long-term climate prediction. We discuss in Sec.~\ref{sec:reservoir} our implementation of reservoir computing using a recurrent neural network (RNN), describe the process of training in Sec.~\ref{sec:training}, and discuss different regularization techniques, whose performance we will test, in Sec.~\ref{sec:regularization}. We then describe our LMNT regularization technique in Sec.~\ref{sec:lmnt}. In Sec.~\ref{sec:prediction}, we discuss how we will produce and evaluate short and long-term predictions using each of the regularization techniques described. Our results using training data from a chaotic test system, the Kuramoto-Sivashinsky equation, are discussed in Sec.~\ref{sec:results}. 
We discuss alternative implementations of LMNT and give concluding remarks in Sec.~\ref{sec:conclusion}.
\subsection{Additional Background: ML Model Robustness to Perturbations}\label{sec:robustness}
A number of techniques have been used to improve ML model ``robustness''  -- insensitivity to small spurious changes -- for input/output situations analogous to Fig.~\ref{fig:basicML}(a). Since this relates to the subject of this article, we discuss some of this past background. These techniques include regularization, such as Tikhonov regularization~\cite{tikhonov_solutions_1977} (also known as ridge regression), LASSO regression~\cite{tibshirani_regression_1996}, and Jacobian regularization~\cite{hoffman_robust_2019}. Training using a dropout scheme~\cite{srivastava_dropout_2014} is another method which introduces random noise to improve model robustness. In RNN models, teacher forcing~\cite{kolen_field_2001}, the more sophisticated professor forcing~\cite{lamb_professor_2016}, or the online weight adjustment technique known as FORCE learning~\cite{sussillo_generating_2009} are used to encourage the dynamics during the closed-loop prediction to more closely resemble the dynamics during training. Noise may also be added to an RNN model's internal state to encourage stability~\cite{lim_noisy_2021}. In the case where an ML prediction model runs in a closed-loop configuration (as in our Fig.~\ref{fig:basicML}(b)), using a cost function that incorporates multiple feedback loops, which can be done simultaneously, as in Ref.~\cite{rasp_weatherbench_2020}, or iteratively, as in Ref.~\cite{pathak_fourcastnet_2022} (referred to as ``fine-tuning''), may also improve model stability. In addition, hybrid methods which combine machine learning with a science-based model can also improve model stability and climate replication~\cite{wikner_combining_2020, arcomano_hybrid_2022}. Many of these regularization techniques introduce additional model ``hyperparameters'' -- algorithmic parameters that are not optimized within the training procedure -- to obtain optimal performance. Tuning hyperparameters can often be computationally expensive, even when \textit{full} optimization is not attempted.

\section{Methods}\label{sec:methods}
We consider the case where we have a finite-duration time series of training data consisting of $M$ simultaneous measurements of state variables from our unknown dynamical system, $\vv{u}(t) = [ u_1(t), u_2(t), \dots, u_M(t)]^\intercal$, obtained from time $t_0$ to time $t_0+t_{\textrm{train}}$ with a sampling time-step of $\Delta t$. We standardize $\vv{u}$ by applying, for each $k$, a linear transformation to the $k^{th}$ measurement time series so that $\vv{u}_k(t)$ has mean $0$ and standard deviation $1$. We assume that the sampling time-step is short compared to the time scale of the unknown dynamical system (e.g., for a chaotic system, the average $e$-fold error growth time, known as the ``Lyapunov time'').

We perform our model training in the open-loop configuration shown in Fig.~\ref{fig:basicML}(a); we input samples from our training time series to our machine learning (ML) model ($\vv{u}_{in}(t)=\vv{u}(t)$), obtaining an $M$-dimensional output $\vv{u}_{out}(t+\Delta t)$. The goal of the training is to adjust the parameters of our ML model so that $\vv{u}_{out}(t+\Delta t) \simeq \vv{u}(t+\Delta t)$ for all $t$ in $t_0\leq t\leq t_0+t_{\textrm{train}}-\Delta t$. Once trained, our model takes an input $\vv{u}(t)$ and outputs a prediction of the dynamical system state at time $t+\Delta t$. We subsequently use the trained model to produce predictions by switching our ML model into the closed-loop configuration shown in Fig.~\ref{fig:basicML}(b). After receiving an initial input up to time $t_{init} \geq t_0 + t_{train}$, the model in this configuration functions autonomously by feeding the model output back into the model as input ($\vv{u}_{in}(t+\Delta t) = \vv{u}_{out}(t+\Delta t)$), shown by the dashed line in Fig.~\ref{fig:basicML}(b). We obtain a prediction for the unknown dynamical system state at time $t_{init}+T\Delta t$, where $T$ is a positive integer, by cycling the model as described $T$ times and recording the final model output as the prediction.
\subsection{Reservoir computing}\label{sec:reservoir}
\begin{figure}[!ht]
    \centering
    \includegraphics[width = \textwidth]{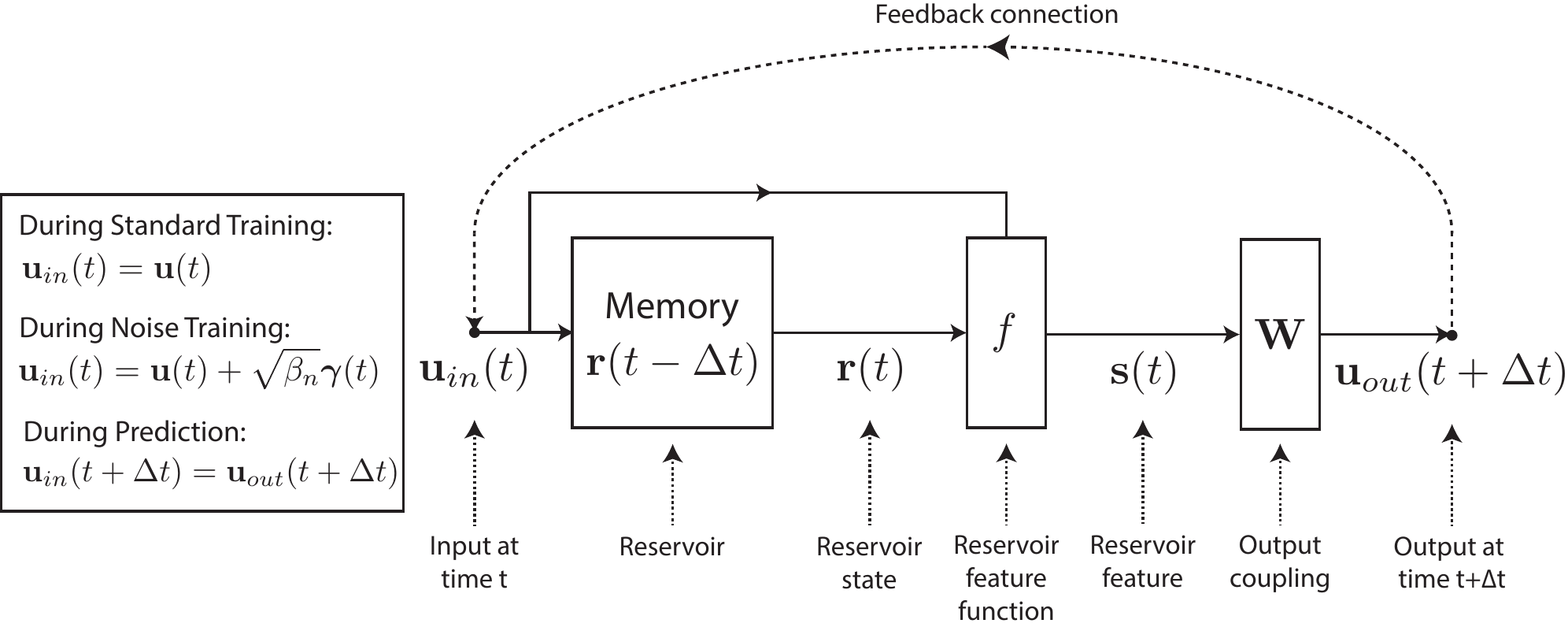}
    \caption{Diagram of our Reservoir Computer Model. The reservoir is trained in the ``open-loop'' training phase (analogous to Fig.~\ref{fig:basicML}(a)). During this phase, the dashed line representing the feedback loop is inactive. Once training is complete, the ``closed-loop'' prediction phase (analogous to Fig.~\ref{fig:basicML}(b)) may begin by activating the output-to-input feedback connection (shown by the dashed line). On the left, we indicate the input during training for the standard training (discussed in Sec.~\ref{sec:training}) and for noise training (discussed in Sec.~\ref{sec:noise_training}), as well as the feedback input during prediction (discussed in Sec.~\ref{sec:prediction}).}
    \label{fig:reservoirdiagram}
\end{figure}
We next describe the particular ML technique used in this article, known as reservoir computing. While we employ reservoir computers for our study, the LMNT approach can be straightforwardly extended to other ML architectures with memory, such as Long-Short Term Memory (LSTM)~\cite{hochreiter_long_1997} or Gated Recurrent Units (GRUs)~\cite{cho_properties_2014}, by approximating the effect of noise added to the input on the output at multiple subsequent times. In reservoir computing, the $M$-dimensional input vector $\vv{u}_{in}(t)$ is coupled by an $N\times M$ input coupling matrix $\vv{B}$ to a high-dimensional reservoir with an $N$-dimensional internal state vector $\vv{r}(t)$. The internal reservoir state at time $t$ depends on both this input and the internal state at time $t-\Delta t$. This internal state gives the reservoir computer a ``memory'' of past system states, which allows it to produce useful predictions even when the $M$-dimensional vector $\vv{u}_{in}(t)$ does not have enough components to represent the full dynamical state of the unknown system to be predicted. This evolving internal state is finally coupled by an output coupling matrix $\vv{W}$ to produce an output $\vv{u}_{out}(t+\Delta t)$. In reservoir computing, only the matrix elements of $\vv{W}$ are adjusted during training, while the values in the input coupling matrix elements and the parameters of the reservoir evolution function are fixed after initialization.

For the reservoir computer implementation used in this article, we use an artificial recurrent neural network with a large number of computational nodes. This network's adjacency matrix, $\vv{A}$, is an $N \times N$ sparse matrix with randomly generated non-zero elements. These non-zero elements represent edges in a directed, weighted graph of $N$ nodes, and the $i^{\text{th}}$ element of the reservoir state vector $\vv{r}(t)$ represents the scalar state of the $i^{\text{th}}$ network node. Values of the non-zero elements of $\vv{A}$ are sampled from a uniform random distribution over the interval $[-1,1]$, and are assigned such that the average number of non-zero elements per row is equal to $\langle d \rangle$, called the ``average in-degree''. For $\vv{A}$ to be sparse, we choose $\langle d \rangle \ll N$. Then, $\vv{A}$ is re-scaled ($\vv{A}\rightarrow[\text{constant}]\times\vv{A}$) such that its ``spectral radius'' $\rho$, the magnitude of the maximum magnitude eigenvalue of $\vv{A}$, is equal to some chosen value. The spectral radius is chosen to be small enough (typically less than 1 suffices) that the reservoir will satisfy the ``echo-state'' property~\cite{jaeger_echo_2001} and large enough that the reservoir has substantial memory. We choose the input coupling $\vv{B}$ as a $N \times M$ matrix with exactly one non-zero value per row so that each reservoir node is coupled to exactly one coordinate of $\vv{u}_{in}(t)$. We choose the location of these non-zero elements such that each input is coupled to an approximately equal number of reservoir nodes. These non-zero elements are sampled from a uniform random distribution over the interval $[-\sigma, \sigma]$; we refer to $\sigma$ as the ``input scaling''.

Given an input $\vv{u}_{in}(t)$ and the previous reservoir internal state $\vv{r}(t-\Delta t)$, the reservoir internal state evolves according to the following equation:
\begin{align}\label{eq:res_evolve}
    \vv{r}(t) = (1-\alpha)\vv{r}(t-\Delta t) + \alpha\tanh(\vv{A}\vv{r}(t-\Delta t) + \vv{B}\vv{u}_{in}(t)+\vv{C}),
\end{align}
where $\tanh$ is applied element-wise. Here, $\vv{C}$ is an $N$-dimensional bias vector with elements sampled from a uniform random distribution over the interval $[-\theta,\theta]$; we refer to $\theta$ as the ``input bias''. The leaking rate, $\alpha$, controls the time scale of the reservoir evolution, and is chosen to be between 0 and 1.

Prior to applying the output coupling $\mathbf{W}$, we compute a ``reservoir feature'', $\vv{s}(t)$, using the reservoir feature function, $\vv{f}(\mathbf{r}(t), \mathbf{u}_{in}(t))$. We choose this function as:
\begin{align}\label{eq:res_fun}
    \vv{s}(t) = \vv{f}(\mathbf{r}(t), \mathbf{u}_{in}(t)) = \begin{bmatrix}
    1 \\
    \mathbf{u}_{in}(t)\\
    \mathbf{r}(t)\\
    \mathbf{r}(t)^2
    \end{bmatrix}.
\end{align}
Here, the $\vv{r}(t)^2$ represents an element-wise power of $2$, while $[\:\vdots\:]$ represents vertical concatenation of the contained elements. The most  essential component of $\vv{s}(t)$ is the reservoir state vector $\vv{r}(t)$, which is a high dimensional vector ($N\gg M$) mapped from the much lower dimensional ($M$) input state space. In addition to $\vv{r}(t)$, the reservoir feature also contains a constant (the $1$), the reservoir input ($\vv{u}_{in}(t)$), and the squared reservoir state ($\vv{r}(t)^2$). We include the reservoir input so that our model need only learn the change from $\vv{u}(t)$ to $\vv{u}(t+\Delta t)$. Finally, we find empirically, as in Ref.~\cite{pathak_model-free_2018}, that including the squared reservoir state improves the accuracy of our forecasts.

The parameters used to generate the reservoir computer ($N$, $\langle d \rangle$, $\rho$, $\sigma$, $\theta$, and $\alpha$) and the regularization parameters used during training (which we will describe in Secs.~\ref{sec:training}-\ref{sec:lmnt}) are ``hyperparameters'' (see Sec.~\ref{sec:robustness}). To obtain an accurate short-term prediction and a long-term prediction with a climate similar to that of the true system, it is necessary to carefully choose the values of these hyperparameters. In reservoir computing, modifying the hyperparameters used to generate a trained reservoir computer will require that one re-perform the training before the reservoir can again be used for prediction. Regularization parameters for some regularization types, however, can be changed without having to re-perform all of the steps in the reservoir training. We will discuss this more in Sec.~\ref{sec:regularization}.

We next describe the reservoir training process and loss function for the different types of regularization we will discuss in this article.
\subsection{Training with Regularization}\label{sec:training}
The goal of the training is to determine the $M\times(1+M+2N)$ output coupling matrix $\vv{W}$ such that $\vv{W}\vv{s}(t) \approx \vv{u}(t+\Delta t)$ when $\vv{u}(t)$ is generated by the unknown dynamical system. Training is accomplished with the reservoir in the ``open-loop'' configuration with no output feedback present. Our training data consists of measurements obtained at $T_{sync}+T_{train}+1$ equally-spaced values:
\begin{align}
\begin{gathered}
    \big\{\vv{u}(t)\big\} = \big\{\vv{u}(0), \vv{u}(\Delta t), \dots, \\
    \vv{u}((T_{sync} + T_{train} - 1)\Delta t), \vv{u}((T_{sync} +T_{train})\Delta t)\big\}.
    \end{gathered}
\end{align}
Each of the $M$ components of $\vv{u}(t)$ has been standardized such that its mean value and standard deviation over the $T_{train}+T_{sync}+1$ time steps are $0$ and $1$, respectively. Here, $T_{train}$ represents the number of training samples, while, as explained subsequently, $T_{sync}$ represents the number of synchronization samples. We begin training by initializing the reservoir so that $\vv{r}(-\Delta t) = \vv{0}$. We then input $\vv{u}_{in}(0) = \vv{u}(0)$, computing the evolved reservoir state $\vv{r}(0)$ and recording the reservoir feature vector $\vv{s}(0)$. Iterating from $t=\Delta t$ to $t = (T_{sync} + T_{train} - 1)\Delta t$, we perform this process, obtaining a set of reservoir feature vectors from $t=0$ to $t = (T_{sync} + T_{train}-1)\Delta t$. We can express the evolution of the reservoir feature vector during training using the open-loop reservoir evolution function, $\vv{g}_o(\vv{s}(t-\Delta t), \vv{u}(t))$, as follows:
\begin{align}
\label{eq:open_loop_1}
    \vv{s}(t) = \: \vv{g}_o(\vv{s}(t-\Delta t), \vv{u}(t))
    = \begin{bmatrix}1\\ \vv{u}(t)\\ \vv{g}_r(\vv{s}(t-\Delta t),\vv{u}(t))\\ \big(\vv{g}_r(\vv{s}(t-\Delta t),\vv{u}(t))\big)^2
    \end{bmatrix},
\end{align}
where
\begin{align}\label{eq:open_loop_2}
\begin{gathered}
\vv{g}_r(\vv{s}(t-\Delta t),\vv{u}(t)) =  (1-\alpha)\begin{bmatrix} \vv{0}_{N\times(1+M)},\:\vv{I}_{N\times N},\: \vv{0}_{N\times N}
\end{bmatrix}\vv{s}(t-\Delta t)\\
+ \alpha \tanh \begin{pmatrix}\begin{bmatrix} \vv{0}_{N\times (1+M)},\:\vv{A},\:\vv{0}_{N\times N}
\end{bmatrix} \vv{s}(t-\Delta t) + \vv{B}\vv{u}(t) + \vv{C}
\end{pmatrix}.
\end{gathered}
\end{align}
In Eq.~\ref{eq:open_loop_2}, $[\dots]$ denotes horizontal concatenation. To ensure that the reservoir feature vectors have minimal dependence on the initial reservoir state, we do not use the first $T_{sync}$ reservoir feature vectors to train the reservoir. The first $T_{sync}$ inputs are only used to ``synchronize'' the reservoir state $\vv{r}(t)$ to the unknown dynamical system trajectory~\cite{pecora_synchronization_2015}. For a reservoir computer with typical hyperparameter values, $T_{sync}$ may be set such that $T_{sync} \ll T_{train}$. We thus obtain $T_{train}$ dynamical system state inputs to the synchronized reservoir, reservoir internal states, and reservoir feature vectors to be used for the training, which we will denote as $\{\vv{u}_{j}\}$, $\{\vv{r}_j\}$, and $\{\vv{s}_j\}$. Here, $0\leq j \leq T_{train}-1$ and sample $j$ is obtained at time $t=(T_{sync}+j)\Delta t$. We additionally introduce the target time series data, $\{\vv{v}_j\}$, which contains the unknown dynamical system states $\vv{u}(T_{sync}+j+1)$ that we desire to approximate at each time index $j$.

We train $\vv{W}$ such that $\vv{W}\vv{s}_j \approx \vv{v}_j$ by minimizing the following regularized least-squares loss function, which, for our purposes, has the form:
\begin{align}\label{eq:res_train}
    \ell(\vv{W}) = \frac{1}{T_{train}}\sum_{j=0}^{T_{train}-1} \lVert \mathbf{W}\mathbf{s}_j - \vv{v}_j \rVert^2_2 + \sum_i\beta_i\text{Tr}\;(\vv{W}\vv{R}_i\vv{W}^\intercal).
\end{align}
In Eq.~\ref{eq:res_train} and in future equations, $\text{Tr}(\dots)$ denotes the trace of a matrix, and $\lVert\dots\rVert_2$ is the Euclidean norm. Here, $\vv{R}_i$ is a regularization matrix, which we denote with an index $i$, with an associated tunable regularization parameter $\beta_i$. Depending on the type of regularization used, $\vv{R}_i$ might or might not depend on $\{\vv{u}_{j}\}$ and $\{\vv{s}_j\}$. We will discuss the different types of regularization used in this article in Secs.~\ref{sec:regularization}$-$\ref{sec:lmnt}. For all regularization types used, we will minimize this cost function using the ``matrix solution'', as follows. We first form the matrices $\vv{S}$ and $\vv{V}$, where the $j^{th}$ columns of $\vv{S}$ and $\vv{V}$ are $\vv{s}_j$ and $\vv{v}_j$, respectively. We then determine a matrix $\vv{W}$ which solves the following linear system obtained by setting the derivative of Eq.~\ref{eq:res_train} with respect to $\vv{W}$ to zero:
\begin{align}\label{eq:train_system_reg}
    \vv{W}\bigg(\frac{1}{T_{train}}\vv{S}\vv{S}^\intercal+\sum_i\beta_i\vv{R}_i\bigg)=\frac{1}{T_{train}}\vv{V}\vv{S}^\intercal.
\end{align}
While all of the regularization methods which we use in this article may be incorporated into the matrix solution, we note that this is not the case for all regularization types (e.g., LASSO regularization~\cite{tibshirani_regression_1996}). In such cases, it would be necessary to use a method other than the matrix solution to minimize Eq.~\ref{eq:res_train} (e.g., proximal gradient methods for LASSO~\cite{daubechies_iterative_2004}). We solve for $\vv{W}$ in Eq.~\ref{eq:train_system_reg} using the \texttt{gesv} function in LAPACK~\cite{laug}, implemented via \texttt{numpy.linalg.solve}~\cite{harris2020array} in Python.

\subsection{Regularization}\label{sec:regularization}
We now discuss the regularization techniques to improve an ML model's climate stability that we will test in this article. Each of these techniques adds a term to the loss function in the form of Eq.~\ref{eq:res_train}, except for noise training, which replaces the vectors $\vv{s}_j$ in Eq.~\ref{eq:res_train} with perturbed vectors $\tilde{\vv{s}}_j$.
\subsubsection{Tikhonov Regularization}\label{sec:tikhonov}
Tikhonov regularization~\cite{tikhonov_solutions_1977}, also known as ``ridge regression'', places a penalty on the Frobenius (element-wise $L^2$) norm of $\vv{W}$ to prevent model over-fitting. In this article, we consider Tikhonov regularization using a single scalar regularization parameter, $\beta_T$; however, one may generally use a matrix of regularization parameters to penalize different elements of $\vv{W}$ to different degrees. The Tikhonov regularization function is
\begin{align}\label{eq:tikhonov_reg_fun}
    \beta_T\lVert \vv{W} \rVert ^2_F=\beta_T\text{Tr}\;(\vv{W}\vv{W}^\intercal)
\end{align}
where $\lVert \dots\rVert_F$ is the Frobenius matrix norm, i.e., the square root of the sum of the squares of all the matrix elements. For use in the last term on the right-hand side of Eq.~\ref{eq:res_train}, we define a Tikhonov regularization matrix as
\begin{align}\label{eq:tikhonov_reg_mat}
    \vv{R}_T = \vv{I}_{(1+M+2N) \times (1+M+2N)},
\end{align}
and express $\text{Tr}(\vv{W}\vv{W}^\intercal)$ as $\text{Tr}(\vv{W}\vv{R}_T\vv{W}^\intercal)$.
\subsubsection{Jacobian Regularization}\label{sec:jacobian}
Jacobian regularization penalizes an ML model's input-output Jacobian matrix to promote model robustness with respect to input perturbations \cite{hoffman_robust_2019}. For reservoir computing, we use the Jacobian matrix of $\vv{u}_{out}(t+\Delta t)=\vv{W}\vv{s}(t)$ with respect to $\vv{u}_{in}(t)$ (see Fig.~\ref{fig:reservoirdiagram}). For the open-loop reservoir evolution function $\vv{g}_o(\vv{s}(t-\Delta t, \vv{u}(t))$ and an input $\vv{u}_j$, this Jacobian is $\vv{W}\bb{\nabla}_\vv{u}\vv{g}_o(\vv{s}_{j-1}, \vv{u}_j)$, where by $\bb{\nabla}_\vv{x}\vv{f}(\vv{y}_j,\vv{x}_j)$ we mean the Jacobian of $\vv{f}(\vv{y},\vv{x})$ with respect to $\vv{x}$ evaluated at $\vv{y}=\vv{y}_j$ and $\vv{x}=\vv{x}_j$. The resulting Jacobian regularization function is
\begin{align}\label{eq:jacobian_reg_fun}
    \frac{\beta_J}{T_{train}-1}\sum_{j=1}^{T_{train}-1}\lVert\vv{W}\bb{\nabla}_\vv{u}\vv{g}_o(\vv{s}_{j-1}, \vv{u}_j)\rVert^2_F = \beta_J\text{Tr}\;(\vv{W}\vv{R}_{\vv{J}}\vv{W}^\intercal),
\end{align}
where the Jacobian regularization matrix is
\begin{align}\label{eq:jac_reg}
    \vv{R}_\vv{J}= \frac{1}{T_{train}-1}\sum_{j=1}^{T_{train}-1}\bb{\nabla}_\vv{u}\vv{g}_o(\vv{s}_{j-1}, \vv{u}_j)\bb{\nabla}_\vv{u}\vv{g}_o(\vv{s}_{j-1}, \vv{u}_j)^\intercal.
\end{align}
We have introduced a normalizing factor of $1/(T_{train}-1)$ so that the scaling of $\vv{R}_{\vv{J}}$ is approximately independent of $T_{train}$. We compute $\bb{\nabla}_\vv{u}\vv{g}_o(\vv{s}_{j-1}, \vv{u}_j)$ from Eqs.~\ref{eq:open_loop_1}~and~\ref{eq:open_loop_2} as follows.
\begin{align}
\begin{split}\label{eq:jacobian}
\bb{\nabla}_\vv{u}\vv{g}_o(\vv{s}_{j-1}, \vv{u}_j)=&\begin{bmatrix}
    0 \\
    \vv{I}_{M\times M} \\
    \alpha\:\di{\vv{h}(\vv{r}_{j-1},\vv{u}_{j})}\vv{B} \\
    \alpha\:\di{2\vv{r}_{j}}\:\di{\vv{h}(\vv{r}_{j-1}, \vv{u}_{j})}\vv{B}
    \end{bmatrix},\quad\textrm{where}
\end{split}\\
\begin{split}\label{eq:diag}
    \di{\vv{d}} = &\begin{bmatrix}
    d_1 & & 0\\
    & \ddots &\\
    0 & & d_N
    \end{bmatrix},\quad\textrm{and}
\end{split}\\
\begin{split}\label{eq:res_derivative}
    \vv{h}(\vv{r}_{j-1},\vv{u}_{j}) =&\text{ sech}^{2}\:(\vv{A}\vv{r}_{j-1}+\vv{B}\vv{u}_{j}+\vv{C}),
\end{split}
\end{align}
where the $\text{sech}^2$ in Eq.~\ref{eq:res_derivative} is the derivative of $\tanh$. We note that Eq.~\ref{eq:jac_reg} depends on the reservoir model parameters ($\vv{A}$, $\vv{B}$, $\vv{C}$, and $\alpha$) and the internal reservoir state, $\vv{r}_{j-1}$, when the reservoir receives the input $\vv{u}_{j}$.

\subsubsection{Noise Training}\label{sec:noise_training}
Similar to our motivation for using Jacobian regularization, we can use noise added to the reservoir input to encourage the reservoir output to be insensitive to perturbations of the input. In noise training, we add a scaled noise vector $\sqrt{\beta_N}\bb{\gamma}(t)$ to each input to the reservoir during training:
\begin{align}
    \vv{u}_{in}(t) = \vv{u}(t) + \sqrt{\beta_N}\bb{\gamma}(t),
\end{align}
where $\bb{\gamma}(t)$ is the noise vector and $\beta_N$ is the noise variance. We generate the set of noise vectors, $\{\bb{\gamma}_j\}$, to be added during training for each of the $M$ components of $\bb{\gamma}(t)$ at sample time $t$, sampling independently from a normal distribution with mean $0$ and standard deviation $1$. Following the process described earlier in this section, we obtain the noisy reservoir features, denoted by $\{\tilde{\vv{s}}_j\}$. We then train $\vv{W}$ such that $\vv{W}\tilde{\vv{s}}_j\approx \vv{v}_j$ by minimizing the noisy loss function:
\begin{align}\label{eq:noisy_loss}
    \ell(\vv{W}) = \frac{1}{T_{train}}\sum_{j=1}^{T_{train}}\lVert \vv{W}\tilde{\vv{s}}_j-\vv{v}_j\rVert^2_2+\beta_T\text{Tr}\:(\vv{WR}_T\vv{W}^\intercal),
\end{align}
where we include a Tikhonov regularization term as well. We minimize Eq.~\ref{eq:noisy_loss} as we did Eq.~\ref{eq:res_train}: by constructing a matrix $\tilde{\vv{S}}$, where the $j^{th}$ column of $\tilde{\vv{S}}$ is $\tilde{\vv{s}}_j$, and solving the following linear system:
\begin{align}\label{eq:noisy_matrix_sol}
    \vv{W}\bigg(\frac{1}{T_{train}}\tilde{\vv{S}}\tilde{\vv{S}}^\intercal+\beta_T\vv{R}_T\bigg)=\frac{1}{T_{train}}\vv{V}\tilde{\vv{S}}^\intercal.
\end{align}
We note that while we can solve Eq.~\ref{eq:noisy_matrix_sol} using the same method used for Eq.~\ref{eq:train_system_reg}, changing the noise scaling $\beta_N$ requires us to re-compute the matrix $\tilde{\vv{S}}$ before we can solve Eq.~\ref{eq:noisy_matrix_sol} again. By contrast, changing the Tikhonov or Jacobian regularization parameter only requires that we re-scale the already-determined regularization matrices before we can solve Eq.~\ref{eq:train_system_reg} again. This re-computation generally makes tuning the noise scaling more computationally costly than the other regularization parameters.
\subsection{Linearized Multi-Noise Training (LMNT)}\label{sec:lmnt}
We now introduce Lineared Multi-Noise Training (LMNT), pronounced as an initialism. While this training technique is motivated by noise training, it is deterministic and results in a regularization matrix that may be easily re-scaled for efficient regularization parameter tuning (thus avoiding the drawback mentioned at the end of Sec.~\ref{sec:regularization}). To formulate this regularization, we consider computing the noisy reservoir feature vectors as described in Sec.~\ref{sec:noise_training} for $P$ different noise realizations added to the input during training, where $P \gg 1$. That is, we compute the reservoir feature vectors for a particular set of noise vectors $\{\bb{\gamma}_{i,1}\}$; then, beginning the computation again, we compute the reservoir feature vectors for another set of noise vectors $\{\bb{\gamma}_{i,2}\}$; and so on, up to iteration $P$. We denote these features as $\{\tilde{\vv{s}}_{j,p}\}$, where the index $p=1,2,\dots,P$ denotes the noise realization used. The full least-squares loss function is
\begin{align}\label{eq:multi_noise_loss}
    \frac{1}{P}\ell(\vv{W}) = \frac{1}{P}\sum_{p=1}^{P}\frac{1}{T_{train}}\sum_{j=0}^{T_{train}-1}\lVert \vv{W}\tilde{\vv{s}}_{j,p}-\vv{v}_j\rVert^2_2+\beta_T\lVert \vv{W}\rVert^2_F,
\end{align}
where we have added a factor of $1/P$ to normalize the loss function to have values similar to Eq.~\ref{eq:res_train}, and we again include a Tikhonov regularization term. We perform a bias-variance decomposition \cite{james_introduction_2014} on the first term in the loss function, obtaining:
\begin{align}\label{eq:variance_decomp}
    \begin{split}
    \frac{1}{P}\sum_{p=1}^{P}\frac{1}{T_{train}}\sum_{j=0}^{T_{train}-1}\lVert \vv{W}\tilde{\vv{s}}_{j,p}-\vv{v}_j\rVert^2_2 = \\ \underbrace{\frac{1}{T_{train}}\sum_{j=0}^{T_{train}-1}\lVert \vv{W}\bar{\vv{s}}_{j}-\vv{v}_j\rVert^2_2}_\text{\clap{bias~}}+\underbrace{\frac{1}{P}\sum_{p=1}^{P}\frac{1}{T_{train}}\sum_{j=0}^{T_{train}-1}\lVert \vv{W}\vv{q}_{j,p}\rVert^2_2}_\text{\clap{~variance}}
    \end{split}
\end{align}
In Eq.~\ref{eq:variance_decomp}, $\bar{\vv{s}}_j$ is the mean of the reservoir feature vector computed over $P$ noise realizations for time index $j$, while $\vv{q}_{j,p}= \tilde{\vv{s}}_{j,p} - \bar{\vv{s}}_j$ is the deviation of $\tilde{\vv{s}}_{j,p}$ from $\bar{\vv{s}}_j$.


We next approximate Eq.~\ref{eq:variance_decomp} by assuming that: (a) $P\rightarrow\infty$; (b) $\beta_N$ is small, allowing us to approximate to linear order in $\sqrt{\beta_N}$; and (c) due the reservoir computer's decaying memory, we can consider only the $K$ most-recent additions of noise prior to time index $j$, where $K\le T_{sync}$. Considering assumption (a), we can write the variance term in Eq.~\ref{eq:variance_decomp} as:
\begin{align}\label{eq:q_covariance}
\begin{split}
    &\lim_{P \to \infty}\frac{1}{P}\sum_{p=1}^{P}\frac{1}{T_{train}}\sum_{j=0}^{T_{train}-1}\lVert \vv{W}\vv{q}_{j,p}\rVert^2_2\\
    =&\lim_{P \to \infty}\frac{1}{T_{train}}\sum_{j=0}^{T_{train}-1}\text{Tr}\bigg[\vv{W}\Big(\frac{1}{P}\sum_{p=1}^P\vv{q}_{j,p}\vv{q}^\intercal_{j,p}\Big)\vv{W}^\intercal\bigg]\\
    =&\frac{1}{T_{train}}\sum_{j=0}^{T_{train}-1}\text{Tr}[\vv{W}\bb{\Sigma}_j\vv{W}^\intercal ],
\end{split}
\end{align} where $\bb{\Sigma}_j$ is the covariance of $\vv{q}_{j,p}$, $\boldsymbol{\Sigma}_j = \lim_{P \to \infty} P^{-1}\sum_{p=1}^P\vv{q}_{j,p}\vv{q}_{j,p}^\intercal$. Next considering assumptions (b) and (c), we write $\tilde{\vv{s}}_{j,p}$ as:
\begin{align}\label{eq:approx_s}
    \tilde{\vv{s}}_{j,p} \approx \vv{s}_j + \sqrt{\beta_N}\sum_{k=j-K+1}^j\bb{\nabla}_\vv{u}(j,k)\bb{\gamma}_{k,p}+\mathcal{O}(\beta_N),
\end{align}
where $\vv{s}_j$ is the feature vector for noiseless training and $\bb{\nabla}_\vv{u}(j,k)$ is the Jacobian of $\vv{s}_j$ with respect to $\vv{u}_k$, where $k \leq j$:
\begin{align}\label{eq:lmnt_jacobian}
\begin{gathered}
    \bb{\nabla}_\vv{u}(j,k) = \bb{\nabla}_\vv{s}\vv{g}_o(\vv{s}_{j-1},\vv{u}_j)\bb{\nabla}_\vv{s}\vv{g}_o(\vv{s}_{j-2},\vv{u}_{j-1})\dots\\\bb{\nabla}_\vv{s}\vv{g}_o(\vv{s}_{k},\vv{u}_{k+1})\bb{\nabla}_\vv{u}\vv{g}_o(\vv{s}_{k-1},\vv{u}_{k}).
\end{gathered}
\end{align}
Then, since $\bb{\gamma}_{k,p}$ is chosen from a distribution with mean zero,
\begin{align}\label{eq:s_bar}
    \lim_{P\rightarrow \infty}\bar{\vv{s}}_j=\vv{s}_j+\mathcal{O}(\beta_N),
\end{align}
and for large P,
\begin{align}\label{eq:approx_q}
    \vv{q}_{j,p} \approx \sqrt{\beta_N} \sum_{k=j-K+1}^j\bb{\nabla}_\vv{u}(j,k)\bb{\gamma}_{k,p}+\mathcal{O}(\beta_N).
\end{align}

We use Eq.~\ref{eq:s_bar} to approximate the bias term in Eq.~\ref{eq:variance_decomp}, and Eq.~\ref{eq:approx_q} to approximate the variance term. We assume that the $\mathcal{O}(\beta_N)$ term in Eq.~\ref{eq:s_bar}, which depends on $j$, has fluctuations that are independent of the fluctuations in $\vv{W}\vv{s}_j-\vv{v}_j$, and that the mean over $j$ of $\vv{W}\vv{s}_j-\vv{v}_j$ is approximately zero, as it should be for least-squares fitting. Thus we assume that changing $\bar{\vv{s}}_j$ to $\vv{s}_j$ in Eq.~\ref{eq:variance_decomp} makes a change that is small compared to $\beta_N$, due to the cancellation of the $\mathcal{O}(\beta_N)$ terms in the expression $\lVert \vv{W}\bar{\vv{s}}_j - \vv{v}_j\rVert_2^2 - \lVert \vv{W}\vv{s}_j - \vv{v}_j\rVert_2^2 \approx 2\vv{W}\mathcal{O}(\beta_N)(\vv{W}\vv{s}_j - \vv{v}_j)^\intercal$ after summing over $j$.

Next, since the noise vectors $\bb{\gamma}_{j,p}$ are independent of each other and each has covariance $\vv{I}$, we approximate $\bb{\Sigma}_j$ in Eq.~\ref{eq:q_covariance} as
\begin{align}\label{eq:lmnt_sigma}
    \bb{\Sigma}_j = \sum_{k=j-K+1}^j\bb{\nabla}_\vv{u}(j,k)\big(\beta_N \vv{I}\big)\bb{\nabla}_\vv{u}(j,k)^\intercal = \beta_N \sum_{k=j-K+1}^j\bb{\nabla}_\vv{u}(j,k)\bb{\nabla}_\vv{u}(j,k)^\intercal.
\end{align}
Combining Eqs.~\ref{eq:q_covariance} and \ref{eq:lmnt_jacobian}, we obtain the LMNT regularization function:
\begin{align}
\beta_L\frac{1}{T_{train}-K}\sum_{j=K}^{T_{train}-1}\sum_{k=j-K+1}^{j}\lVert \vv{W}\bb{\nabla}_\vv{u}(j,k)\rVert^2_F,
\end{align}
where we have replaced $\beta_N$ by $\beta_L$. We restrict to $j \geq k$ so that $k\geq 1$ and, thus, $k-1\geq 0$ in Eq.~\ref{eq:lmnt_jacobian}. In Eq.~\ref{eq:lmnt_jacobian}, $\vv{u}_{k}$ and $\vv{s}_{k}$ are computed without any addition of noise, making this regularization deterministic. We use Eq.~\ref{eq:jacobian} for $\bb{\nabla}_\vv{u}\vv{g}_o$, and we compute $\bb{\nabla}_\vv{s}\vv{g}_o$ using the open-loop reservoir feature evolution described by Eq.~\ref{eq:open_loop_1} and shown in Fig.~\ref{fig:reservoirdiagram}:
\begin{align}
\begin{split}
    \bb{\nabla}_{\vv{s}}\vv{g}_o(\vv{s}_{j-1},\vv{u}_j) = {}&\begin{bmatrix}
    \mathbf{0}_{(1+M)\times(1+M+2N)}\\
    \bb{\nabla}_{\vv{s}}\vv{g}_r(\vv{s}_{j-1},\vv{u}_j)\\
    \di{2\vv{r}_{j}}\:\bb{\nabla}_{\vv{s}}\vv{g}_r(\vv{s}_{j-1},\vv{u}_j)
    \end{bmatrix},\quad \textrm{where}
\end{split}\\
\begin{split}
\bb{\nabla}_{\vv{s}}\vv{g}_r(\vv{s}_{j-1},\vv{u}_j) = {}&\alpha\:\di{ \vv{h}(\vv{r}_{j-1}, \vv{u}_{j})}\begin{bmatrix}
\vv{0}_{N\times (1+M)}, & \vv{A}, & \vv{0}_{N \times N}
\end{bmatrix} \\{}&+ \begin{bmatrix}
\vv{0}_{N\times(1+M)}, & (1-\alpha)\vv{I}_{N\times N}, & \vv{0}_{N\times N}
\end{bmatrix}
\end{split}
\end{align}
and where $\di{\dots}$ and $\vv{h}(\vv{r}_{j-1},\vv{u}_{j})$ are the same as in Eqs.~\ref{eq:diag}~and~\ref{eq:res_derivative}, respectively. When computed using the inputs and internal reservoir states during training, the resulting regularization matrix is:
\begin{align}\label{eq:lmnt}
    \vv{R}_L = \frac{1}{T_{train}-K}\sum_{j=K}^{T_{train}-1}\Bigg[\sum_{k=j-K+1}^j\bb{\nabla}_\vv{u}(j,k)\bb{\nabla}_\vv{u}(j,k)^\intercal\Bigg].
\end{align}
In the case where $K=1$, this regularization matrix is identical to the Jacobian regularization matrix, $\vv{R}_J$. For all results in this article, we will compute the LMNT regularization using $K=4$; we justify this choice of $K$ and show results for other $K$ values in~\ref{app:lmnt_K}. We also note that the LMNT regularization is computed here using approximately the same number of training samples as is used to train the reservoir. In~\ref{app:reduced_lmnt}, we discuss computing the LMNT regularization with a greatly reduced number of training samples.

\subsection{Prediction and Metrics}\label{sec:prediction}
Once we have determined $\vv{W}$, we are ready to begin prediction. Prior to switching on the feedback loop shown in Fig.~\ref{fig:reservoirdiagram}, we reset the internal reservoir state to $\vv{0}$, and then input a short sequence of measurements, $\{\vv{u}_{true}(t)\}$, from the system we intend to predict. More precisely, assume that $\vv{u}_{true}(t)$ is known up to time $T_{init}\Delta t$. To re-synchronize the reservoir to the true system trajectory, we begin at time $ (T_{init}-T_{sync})\Delta t$, inputting $\vv{u}_{in}((T_{init}-T_{sync})\Delta t) = \vv{u}_{true}((T_{init}-T_{sync})\Delta t)$. We iterate this process from $t = (T_{init}-T_{sync})\Delta t$ to $t = T_{init}\Delta t$, recording the resulting reservoir states until we obtain $\vv{r}(T_{init}\Delta t)$. We then compute our prediction for the system state at time $(T_{init}+1)\Delta t$ as $\vv{u}_{out}((T_{init}+1)\Delta t) = \vv{W}\vv{s}(T_{init}\Delta t)$, where $\vv{s}(T_{init}\Delta t) = f(\vv{r}(T_{init}\Delta t), \vv{u}_{true}(T_{init}\Delta t))$ (see Eq.~\ref{eq:res_fun}). We then activate the feedback loop shown in Fig.~\ref{fig:reservoirdiagram} so that $\vv{u}_{in}(t) = \vv{u}_{out}(t)$ for $t\geq(T_{init}+1)\Delta t$. We compute $\vv{u}_{out}((T_{init}+2)\Delta t)=\vv{Ws}((T_{init}+1)\Delta t)$, feed back this prediction as input, and so on, until we have reached our desired prediction time, $t_{pred} = (T_{init} + T_{pred})\Delta t$, where $T_{pred}$ is a positive integer. The closed-loop reservoir dynamics can be expressed as a single evolution function $\vv{s}(t) = \vv{g}_c(\vv{s}(t-\Delta t))$, where

\begin{align}\label{eq:closed_loop}
    \vv{g}_c(\vv{s}(t-\Delta t)) = \begin{bmatrix}1\\ \vv{W}\vv{s}(t-\Delta t)\\ \vv{g}_s(\vv{s}(t-\Delta t))\\ \big(\vv{g}_s(\vv{s}(t-\Delta t))\big)^2
    \end{bmatrix}
\end{align}
and
\begin{align}\label{eq:closed_loop_g}
\begin{split}
\vv{g}_s(\vv{s}(t-\Delta t)) =  \:& (1-\alpha)\begin{bmatrix} \vv{0}_{N\times(1+M)},\:\vv{I}_{N\times N},\: \vv{0}_{N\times N}
\end{bmatrix}\vv{s}(t-\Delta t) \\
&+ \alpha \tanh \begin{pmatrix}
\begin{bmatrix}\vv{0}_{N\times 1},\: \vv{BW},\:\vv{A},\:\vv{0}_{N\times N}
\end{bmatrix} \vv{s}(t-\Delta t)+\vv{C}
\end{pmatrix}.
\end{split}
\end{align}
The right side of Eq.~\ref{eq:closed_loop_g} is obtained by substituting $\vv{u}(t)=\vv{Ws}(t+\Delta t)$ into the right side of Eq.~\ref{eq:open_loop_2}.

When evaluating the performance of a prediction from our machine learning model, we are interested in:
\begin{enumerate}
    \item For what duration of time is the near-term prediction approximately valid? (In other words, how long does the near-term prediction error remain below some chosen threshold?)
    \item Is the long-term climate ``stable''? (In other words, does the ML model prediction remain within approximately the same region of space as the training data, or does it escape to some other region?)
    \begin{itemize}
        \item Due to the chaotic nature of the systems we are interested in, we expect predictions to be ``unstable'' in the sense that the predicted trajectory will exponentially diverge from the true trajectory due to any error in the initial condition, no matter how small the error is in the model. We distinguish this type of instability from climate instability.
    \end{itemize}
    \item If the prediction is stable, are its statistical properties, or ``climate'', similar to that of the unknown dynamical system?
\end{enumerate}
To evaluate each of these criteria, we use the following metrics:
\begin{enumerate}
    \item \textbf{Prediction Valid Time:} The prediction valid time is computed as
    \begin{align}\label{eq:valid_time}
        VT = \min_{T_{init}\Delta t \leq t \leq (T_{init}+T_{pred})\Delta t} \Big\{ t \:\big|\:\frac{\lVert \vv{u}_{out}(t) - \vv{u}_{true}(t)\rVert_2}{\overbar{E}} > \epsilon_{VT} \Big\} - \Delta t.
    \end{align}
    Here, $\epsilon_{VT}$ is the valid time error threshold, and $\overbar{E}$ is the average error between true system states computed from the training data as the mean of $\lVert \vv{u}_j - \vv{u}_k\rVert_2$ over $0\leq j < k \leq T_{train}$. In all of our tests, we choose $\epsilon_{VT} = 0.2$, representing a $20\%$ error in the prediction.
    \item \textbf{Climate Stability:} It is often the case that unstable ML model predictions will diverge exponentially from the region of the true attractor, eventually resulting in numerical overflow in the floating-point computation. These predictions can be easily identified as unstable. However, unstable predictions may also settle in some other region of space instead, necessitating that we use some other metric to characterize if the climate of such predictions is stable. We determine if such a prediction is stable by computing the mean of the normalized ``map error'' over the entire prediction. This metric presumes a diagnostic scenario where we know the true evolution equations for $\vv{u}(t)$ and that $\vv{u}(t)$ is the entire state of the system to be predicted. The map error at time $t$ is the norm of the difference between $\vv{u}_{out}(t)$ and the result of evolving the true evolution equations for time $\Delta t$ from initial condition $\vv{u}_{out}(t-\Delta t)$. The normalized map error is
    \begin{align}
        \epsilon_{map}(t)=\frac{\lVert \vv{u}_{out}(t) - \vv{F}(\vv{u}_{out}(t-\Delta t), t-\Delta t, t)\rVert_2}{\overbar{E}_{map}}.
    \end{align}
    Here, $\vv{F}(\vv{u}(t_0), t_0, t_f)$ is a function that integrates the true evolution equations with an initial condition $\vv{u}(t_0)$ from $t_0$ to $t_f$. We have normalized the map error using the mean error of the persistence forecast computed from the training data,
    \begin{align}
        \overbar{E}_{map} = \overline{\lVert \vv{u}_{j+1} - \vv{u}_j\rVert_2},
    \end{align}
    where the horizontal bar $\overline{(\dots)}$ denotes a mean computed over the training time indices from $j=0$ to $j=T_{train}-1$.

    We denote by $\bar{\epsilon}_{map}$ the mean of $\epsilon_{map}(t)$ for $T_{init}\Delta t \leq t \leq (T_{init} + T_{pred})\Delta t$. We choose a threshold for the mean normalized map error that, if exceeded, characterizes the prediction as unstable. We choose this threshold by first producing predictions using an ensemble of reservoir realizations, training data sets, and testing data sets that are each trained using regularization parameter value from a logarithmic grid of parameter values. We then compute a histogram of the mean map error values from predictions that have not resulted in numerical overflow. In all cases we test, we are able to clearly see a multimodal distribution, with those points with low mean map error corresponding to stable predictions and those with high mean map error corresponding to unstable predictions. We show an example of a histogram of mean map error values in Sec.~\ref{sec:test_results}. From this histogram, we have chosen $\bar{\epsilon}_{map} = 1.0$ as our stability cutoff. To measure the maximum deviation from the true evolution equations during prediction, we also compute the maximum map error for each prediction,
    \begin{align}
        \epsilon_{map}^{max} = \max_{T_{init}\Delta t\leq t \leq (T_{init}+T_{pred})\Delta t} \epsilon_{map}(t).
    \end{align}
    \begin{sloppypar}
    \item \textbf{Climate Similarity:} To compare the climate of the true system and the stable predictions generated from our reservoir computer, we use the power spectral density (PSD), $S_{uu}(f)$. We estimate the PSD of a particular element of $\vv{u}_{true}(t)$ or $\vv{u}_{out}(t)$ using a smoothed periodogram. For a 1-dimensional time series data set $\{u_j\}=\{u_0, u_1, \dots, u_{J-2}, u_{J-1}\}$, where $J=T_{pred}/\Delta t$ and $u_j = u((T_{init}+j+1)\Delta t)$, this estimate is computed using Welch's method~\cite{welch_use_1967}.
    \end{sloppypar}
\end{enumerate}
\section{Results}
\label{sec:results}
\subsection{Kuramoto-Sivashinsky Equation}
The test system that we will use to examine near-term accuracy and long-term climate stability and accuracy is the Kuramoto-Sivashinsky (KS) equation \cite{kuramoto_diffusion-induced_1978, sivashinsky_nonlinear_1977}
with periodic boundary conditions:
\begin{align}\label{eq:ks_true}
    \frac{\partial y(x,t)}{\partial t} + y(x,t)\frac{\partial y(x,t)}{\partial x} + \frac{\partial^2 y(x,t)}{\partial x^2} + \frac{\partial^4 y(x,t)}{\partial x^4}=0,
\end{align}
where $y(x,t) = y(x+L,t)$ for a chosen spatial periodicity length $L$. For particular choices of $L$, the Kuramoto-Sivashinky equation exhibits chaotic dynamics. To obtain the dynamical system states that we will use as our training and testing data, and to evaluate the map error as discussed in Sec.~\ref{sec:prediction}, we simulate Eq.~\ref{eq:ks_true} on a spatial grid consisting of $64$ grid points equally-spaced at intervals of $\Delta x = L/64$ using the ETRK4 method~\cite{cox_exponential_2002-1, kassam_fourth-order_2005} to integrate the system at a time step of $\Delta t = 0.25$. The resulting discretized system state is:
\begin{align}
    \vv{y}(t) = [y(0,t), y(\Delta x,t), \dots, y((M-1)\Delta x, t)]^\intercal,
\end{align}
where $M = 64$. For each time series used for training and testing, we choose a different random initial condition, where each coordinate of $\vv{u}(0)$ is sampled from a uniform distribution on the interval $[-0.6, 0.6]$, with the spatial mean value of $\vv{u}(0)$ adjusted to be $0$. We then integrate the equation and discard the the states obtained obtained before $t=500$ to avoid the effect of any transient dynamics that are not on the true attractor. Our numerical integrations of Eq.~\ref{eq:ks_true} yield chaotic trajectories and (as they should) preserve the zero spatial average of $\vv{y}(t)$ for $t>0$. We standardize $\vv{y}(t)$ as follows to form $\vv{u}(t)$ that is used for training and testing the reservoir. To each coordinate of $\vv{y}(t)$, we apply a linear transformation so that the resulting coordinate of $\vv{u}(t)$ has mean $0$ and standard deviation $1$ over the training time period.

When discussing the prediction valid time and the prediction spectra, we will do so in units of the Lyapunov time, corresponding to the inverse of the largest positive Lyapunov exponent for typical orbits of the chaotic attractor of Eq.~\ref{eq:ks_true}.  The Lyapunov time is characteristic time over which errors in the true chaotic system will experience an $e$-fold growth. For a periodicity length $L=22$, we have computed the Lyapunov time to be $t_{Lyap}=20.83$ using the Bennetin algorithm \cite{bennetin_lyapunov_1980}.
\begin{sloppypar}
We have also done tests with other systems besides the Kuramoto-Sivashinky equations (e.g., the chaotic Lorenz '63 model \cite{lorenz_deterministic_1963}). We do not report these results here, as they result in conclusions that coincide with what follows from our tests on Eq.~\ref{eq:ks_true}.
\end{sloppypar}
\subsection{Prediction Test Results}\label{sec:test_results}
We now present the result of our predictions test on the Kuramoto-Sivashinky equation. In order to test a scenario where climate stability is challenging, we have used a reservoir with $500$ nodes (see Table~\ref{table:res_hyperparams}). Though it would be computationally feasible to improve stability with a larger reservoir in this case, it might not be in scenarios with higher dimensional dynamics. In summary, we find from our tests that:
\begin{itemize}
    \item Reservoir computers trained with no regularization or only Jacobian regularization produce predictions that are always observed to be unstable and produce predictions that have a very low valid time.
    \item Reservoir computers trained with only Tikhonov regularization produce predictions that are only sometimes observed to be stable, while otherwise producing predictions with a very low valid time.
    \begin{sloppypar}
    \item Reservoir computers trained with Jacobian and Tikhonov regularization, noise training and Tikhonov regularization, and LMNT and Tikhonov regularization are observed to always produce stable predictions if the regularization hyperparameters are chosen large enough. The latter two regularization methods result in the best prediction valid times, as well as the best mean and maximum map errors.
    \end{sloppypar}
    \begin{sloppypar}
    \item Stable predictions from reservoir computers trained with only Tikhonov regularization have an average PSD that appears to match the PSD of the true system climate somewhat well, while predictions from the methods that always produce stable predictions have an average PSD that near-perfectly matches the PSD of the true system climate.
    \end{sloppypar}
    \item From our variation of the regularization parameters, we find the best prediction valid time performance near the boundary between climate stability and partial climate instability, indicating that careful tuning of the regularization parameter value is needed during optimization.
\end{itemize}
\subsubsection{Climate Stability and Valid Time}
\begin{center}
\begin{table}[!ht]
\centering
    \begin{tabular}{|l|l|}
        \hline
        \multicolumn{2}{|c|}{Reservoir Hyperparameters}\\
        \hline\hline
         Number of nodes ($N$) & $500$ \\ \hline
         Average degree ($\langle d \rangle$)& $3$\\ \hline
         Spectral radius ($\rho$) & $0.6$ \\ \hline
         Input weight ($\sigma$) & $0.1$ \\ \hline
         Input bias ($\theta$) & $0.1$ \\ \hline
         Leaking rate ($\alpha$) & $1.0$ \\ \hline
    \end{tabular}
    \caption{Reservoir computer hyperparameter values used for all prediction tests.}
    \label{table:res_hyperparams}
    \end{table}
\end{center}

\begin{center}
\begin{table}[!ht]
    \centering
    \resizebox{\columnwidth}{!}{
    \begin{tabular}{|l|l|}
        \hline
        \multicolumn{2}{|c|}{Training and Prediction Parameters}\\
        \hline \hline
         $T_{sync}$ & $100$ ($1.2\:t_{Lyap}$) \\ \hline
         $T_{train}$& $20,000$ ($240\:t_{Lyap}$) \\ \hline
         $T_{pred}$ & \begin{tabular}{l} Valid Time Tests: $2000$ ($24\:t_{Lyap}$)\\ Climate Stability and Accuracy Tests: $16,000$ ($192\:t_{Lyap}$)\end{tabular} \\ \hline
         Number of Reservoir Ensemble Members &  \begin{tabular}{l} $20$ \end{tabular}\\ \hline
         Number of Training Data Sets & \begin{tabular}{l} $10$ \end{tabular}\\ \hline
         Number of Testing Data Sets & \begin{tabular}{l} Prediction Valid Time Tests: $35$ \\ Climate Stability and Accuracy Tests: $5$\end{tabular}\\ \hline
         Total Number of Predictions & \begin{tabular}{l} Prediction Valid Time Tests: $7000$ \\ Climate Stability and Accuracy Tests: $1000$\end{tabular}\\ \hline
    \end{tabular}
    }
    \caption{Parameters used for all reservoir training and tests.}
    \label{table:testing_params}
    \end{table}
\end{center}

\begin{center}
\begin{table}[!ht]

    \resizebox{\columnwidth}{!}{
    \begin{tabular}{|l||l|}
        \hline
        \multicolumn{2}{|c|}{Regularization Parameter Search Grids}\\\hline
        Regularization Type & Search Grid \\ \hline\hline
        Tikhonov ($\beta_T$)&  0 and $10^{l}$ for $l \in \{-18,-17.5,\dots, -4.5, -4\}$\\ \hline
        Jacobian ($\beta_J$)&  $10^{l}$ for $l \in \{-8, -7.8, \dots,  -4.2, -4\}$ \\ \hline
        Noise Training ($\beta_N$) and LMNT ($\beta_L$)& $10^{l}$ for $l \in \{-8, -7.8, \dots,  -6.2, -6\}$\\ \hline
    \end{tabular}
    }
    \caption{The regularization parameters that are searched over for each reservoir prediction test.}
    \label{table:test_grid}
    \end{table}
\end{center}

\begin{figure}[!ht]
    \centering
    \includegraphics[width = \textwidth]{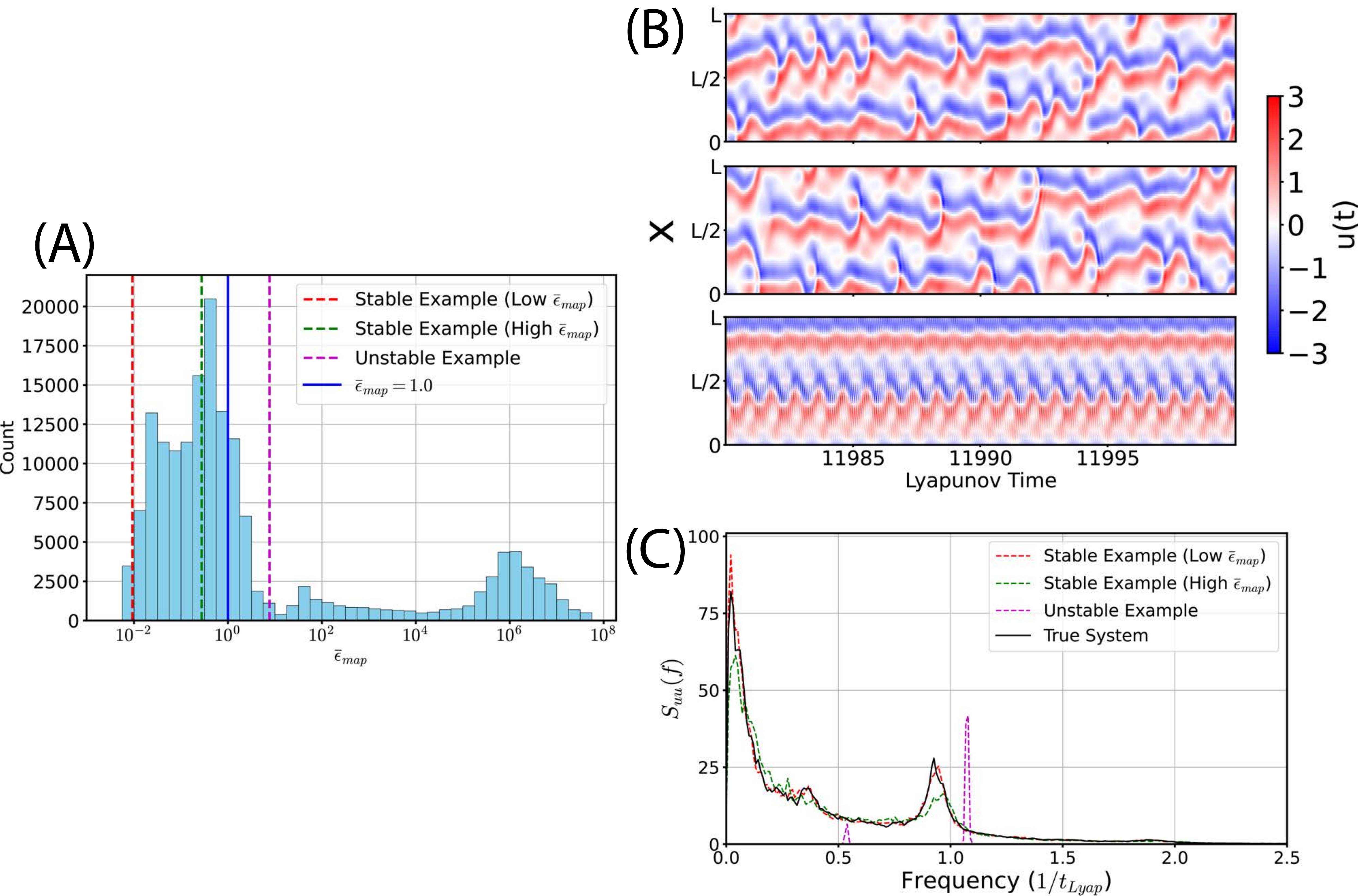}
    \caption{Comparing Predictions with Increasing $\bar{\epsilon}_{map}$ Values. In this figure, panel (A) shows a histogram of the mean map error values of predictions generated from reservoir computers trained with Jacobian and Tikhonov regularization using a logarithmic grid of regularization parameter values (grid given in Table~\ref{table:test_grid}). In this panel, the solid blue line marks the $\bar{\epsilon}_{map}$ climate stability threshold, while the dashed red, green, and magenta lines denote the mean map error value for three example predictions: one is a stable prediction with a low $\bar{\epsilon}_{map}$, one is stable with a high $\bar{\epsilon}_{map}$, and one is unstable, respectively. These predictions are obtained using a long $T_{pred}=1,000,000\;(12,000\:t_{Lyap})$. The dynamics at the end of these predictions are displayed in panel (B), with the stable example prediction with a low $\bar{\epsilon}_{map}$ on the top, the stable example prediction with a high $\bar{\epsilon}_{map}$ in the middle, and the unstable example prediction on the bottom. Panel (C) displays the computed power spectral density of $u_{out,1}(t)$ and $u_{true,1}(t)$ for each of these predictions and the true test data. When computing the power spectral density using Welch's method, we used a window with $2^{13}$ samples $(98.304\: t_{Lyap})$.}
    \label{fig:ks_stab_histogram}
\end{figure}

\begin{center}
\begin{table}[!ht]
    \centering
    \resizebox{\columnwidth}{!}{
    \begin{tabular}{|l||l|l|l|l|l|}
        \hline
        \multicolumn{6}{|c|}{Prediction Test Results}\\
        \hline
         Regularization Type &  \begin{tabular}{l} Regularization\\ Parameters\end{tabular} & \begin{tabular}{l}Fraction \\of Stable \\Predictions \end{tabular}& \begin{tabular}{l} Median \\Valid Time \\ ($t_{Lyap}$) \end{tabular}& \begin{tabular}{l} Median \\$\bar{\epsilon}_{map}$ \end{tabular}& \begin{tabular}{l}Median \\$\epsilon_{map}^{max}$\end{tabular}\\ \hline\hline
         None & N/A & $0/1000$ & $0.05\pm 0.01$ & $\infty$ & $\infty$ \\ \hline
         Jacobian Only & $\beta_J = 10^{-7}$ & $0/1000$ & $0.25\pm 0.01$ & $\infty$ & $\infty$\\ \hline
         Tikhonov Only & $\beta_T = 10^{-6}$ & $565/1000$ & $0.71\pm0.02$ & \begin{tabular}{l} $(6.46\pm 0.22)$\\$\times10^{-1}$\end{tabular} & $5.23\pm 0.31$\\ \hline
         \begin{tabular}{l} Jacobian and \\ Tikhonov \end{tabular}& \begin{tabular}{l} $\beta_J = 10^{-5.4}$ \\ $\beta_T = 10^{-8.5}$  \end{tabular} & $\vv{1000/1000}$ & $2.88\pm 0.02$ &\begin{tabular}{l}$(9.16\pm 0.04)$\\$\times10^{-3}$ \end{tabular}& \begin{tabular}{l}$(5.82\pm 0.06)$\\$\times10^{-2}$ \end{tabular}\\ \hline
         \begin{tabular}{l} Noise Training\\ and Tikhonov \end{tabular}& \begin{tabular}{l} $\beta_N = 10^{-7.4}$ \\ $\beta_T = 10^{-14.5}$ \end{tabular} & $\vv{1000/1000}$ & $\vv{4.24}\bb{\pm} \vv{0.04}$ & \begin{tabular}{l}$\vv{(2.77}\bb{\pm}\vv{ 0.02)}$\\$\bb{\times} \vv{10}^{\vv{-3}}$ \end{tabular}&  \begin{tabular}{l}$\vv{(2.02}\bb{\pm}\vv{0.04)}$\\$\bb{\times} \vv{10}^{\vv{-2}}$\end{tabular}\\ \hline
         \begin{tabular}{l} LMNT ($K=4$) \\ and Tikhonov \end{tabular}& \begin{tabular}{l} $\beta_L = 10^{-7.4}$ \\ $\beta_T = 10^{-16.5}$ \end{tabular} & $\vv{1000/1000}$ & $\vv{4.27}\bb{\pm} \vv{0.04}$ & \begin{tabular}{l}$\vv{(2.75}\bb{\pm}\vv{ 0.02)}$\\$\bb{\times} \vv{10}^{\vv{-3}}$ \end{tabular} & \begin{tabular}{l}$\vv{(2.03}\bb{\pm}\vv{0.04)}$\\$\bb{\times} \vv{10}^{\vv{-2}}$\end{tabular}\\ \hline
    \end{tabular}
    }
    \caption{Reservoir computer prediction results using different types of regularization. Bold text marks the best performance for the corresponding metric. In every case, regularization parameter values are chosen to maximize the fraction of stable predictions produced by an ensemble of reservoirs over an ensemble of training and testing data sets. If multiple regularization parameter values are found to produce the same fraction of stable predictions, then we choose the regularization parameter(s) from that subset that maximize the median prediction valid time. The $\pm$ error bounds indicate the maximum of the upper and lower $95\%$ confidence intervals for the median \cite{conover_practical_1999}. In the case of the valid time, we enlarge this interval slightly to account for the $\Delta t$ discretization of our predictions.}
    \label{table:test_results}
    \end{table}
\end{center}

\begin{figure}[!ht]
    \centering
    \includegraphics[width = 0.9\textwidth]{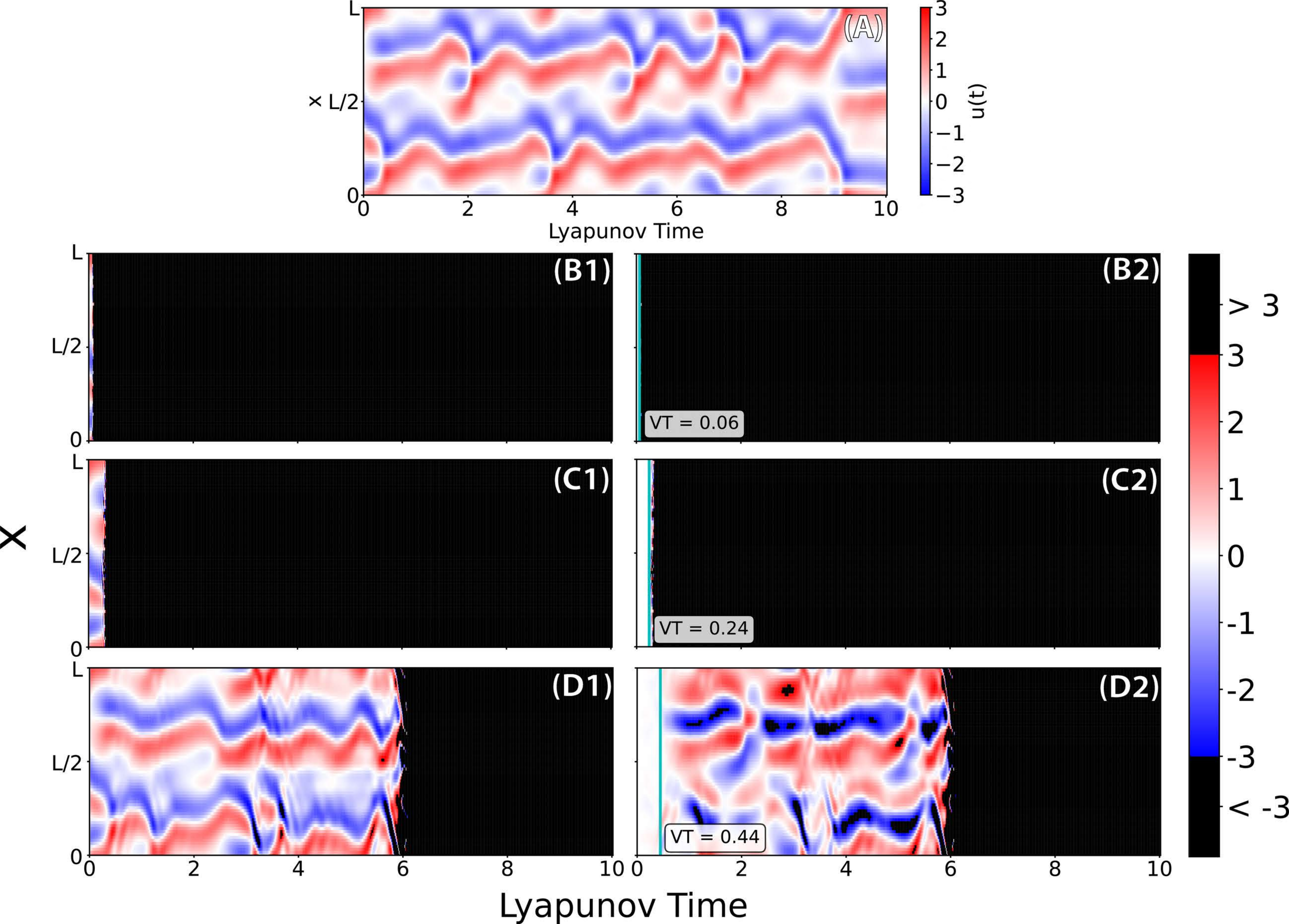}
    \caption{Unstable Predictions of the KS Equation. The top panel (A) shows a testing time series $\vv{u}_{true}(t)$ generated from the true KS equations dynamics, while the bottom panels show results from a reservoir computer trained using no regularization (B1 and B2), Jacobian regularization only (C1 and C2), and Tikhonov regularization only (D1 and D2). Panels on the left show the reservoir prediction, $\vv{u}_{out}(t)$, while those on the right show the difference between the reservoir prediction and the true evolution $\vv{u}_{true}(t)$. In all panels, the horizontal axis denotes the time measured in Lyapunov times based on the largest positive Lyapunov exponent of the true system, while the vertical axis denotes the spatial coordinate. In panel (A) and in the panels on the left, the color denotes the values of $\vv{u}_{true}(t)$ and $\vv{u}_{out}(t)$, respectively, while in panels on the right, the color denotes the difference, $\vv{u}_{out}(t)-\vv{u}_{true}(t)$, between the prediction and true dynamics. An entire prediction at a particular time being colored black indicates that this prediction has become unstable and left the region of the true attractor. In the right panels, the cyan line denotes the corresponding prediction valid time.}
    \label{fig:ks_preds_unstable}
\end{figure}
\begin{figure}[!ht]
    \centering
    \includegraphics[width = 0.9\textwidth]{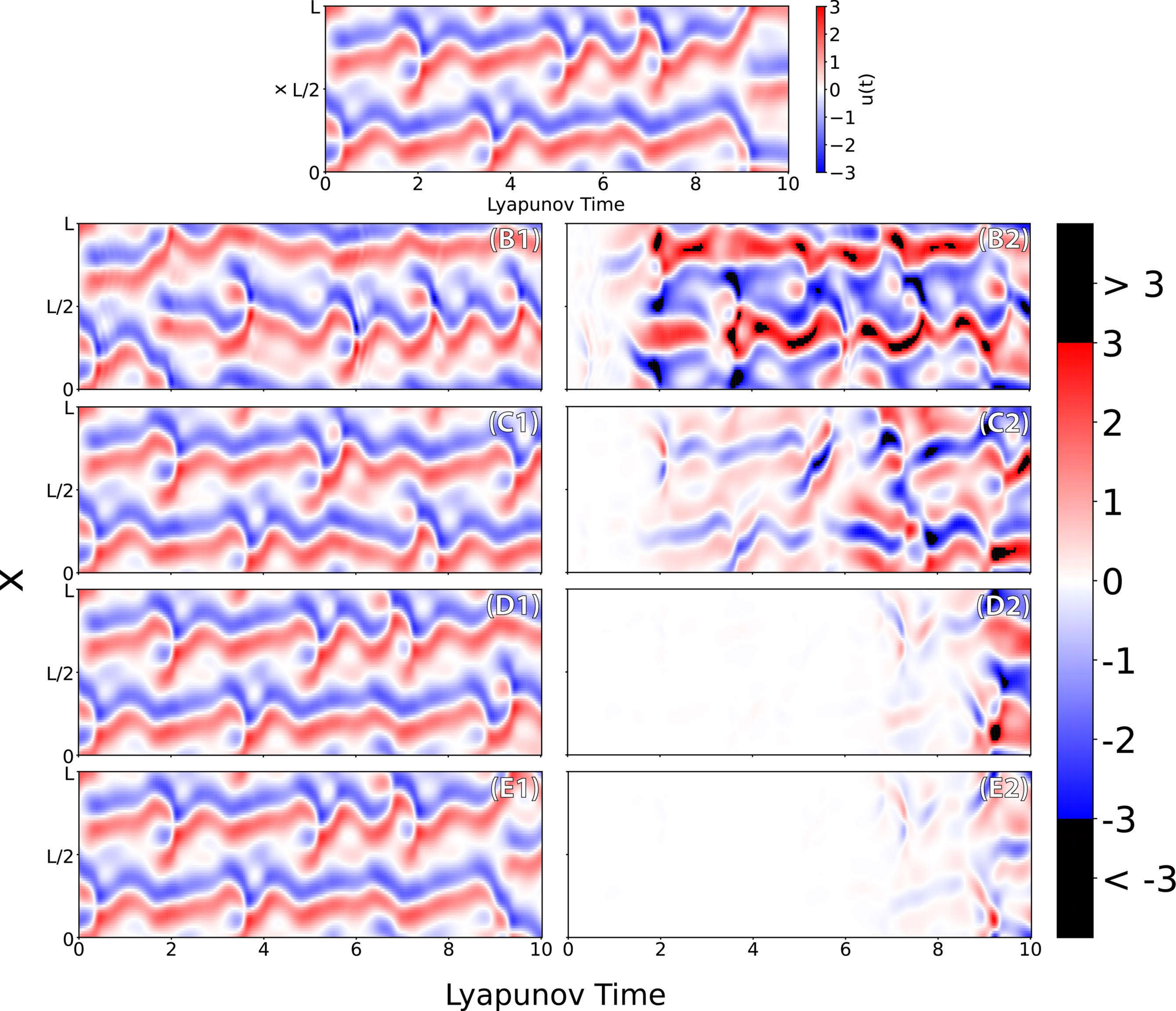}
    \caption{Stable Predictions of the KS Equation. The top panel (A) shows a testing time series $\vv{u}_{true}(t)$ generated from the true KS equations dynamics, while the bottom panels show results from a reservoir computer trained using Tikhonov regularization only (B1 and B2), Jacobian regularization and Tikhonov regularization (C1 and C2), noise training and Tikhonov regularization (D1 and D2), LMNT and Tikhonov regularization (E1 and E2). Panels on the left show the reservoir prediction, $\vv{u}_{out}(t)$, while those on the right show the difference between the reservoir prediction and the true evolution $\vv{u}_{true}(t)$. In all panels, the horizontal axis denotes the time measured in Lyapunov times based on the largest positive Lyapunov exponent of the true system, while the vertical axis denotes the spatial coordinate. In panel (A) and in the panels on the left, the color denotes the values of $\vv{u}_{true}(t)$ and $\vv{u}_{out}(t)$, respectively, while in panels on the right, the color denotes the difference, $\vv{u}_{out}(t)-\vv{u}_{true}(t)$, between the prediction and true dynamics. Black coloring marks where the prediction (left panels) or prediction difference (right panels) is outside the range given by the color bar.}
    \label{fig:ks_preds_stable}
\end{figure}
\begin{figure}[!ht]
    \centering
    \includegraphics[width = 0.9\textwidth]{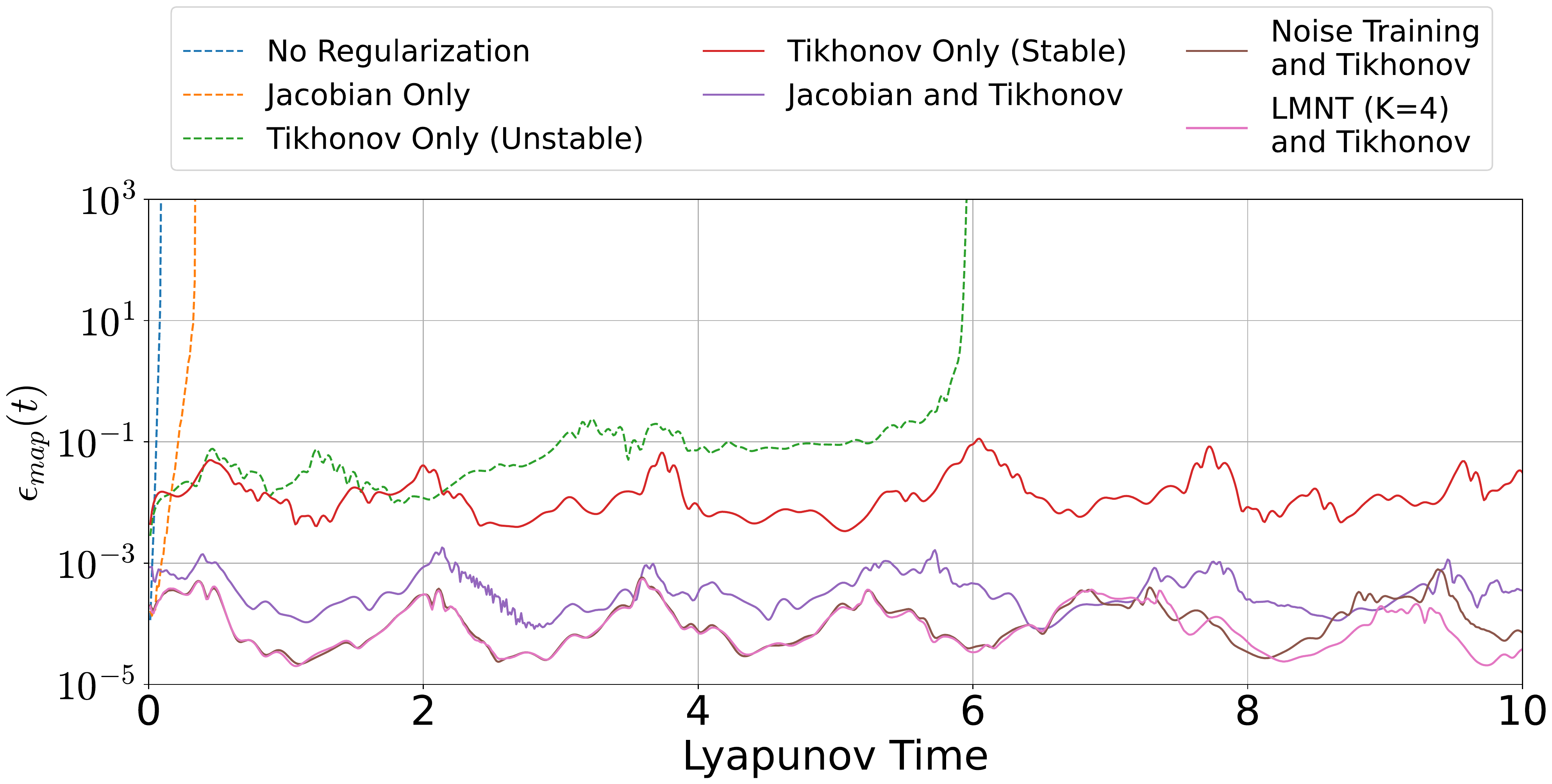}
    \caption{Map Error over Time During Reservoir Predictions. This figure shows the logarithm of the map error $\epsilon_{map}$ during each of the predictions shown in Fig.~\ref{fig:ks_preds_unstable} (dashed lines) and Fig.~\ref{fig:ks_preds_stable} (solid lines).}
    \label{fig:ks_map_error}
\end{figure}

We simulate the KS equation using reservoir computers with the hyperparameters listed in Table~\ref{table:res_hyperparams}. For each of our tests, we train an ensemble of random reservoir realizations, each generated using a different random seed for the input coupling matrix $\vv{B}$ and the reservoir network adjacency matrix $\vv{A}$, using a training time series data set drawn from an ensemble of training data sets generated using Eq.~\ref{eq:ks_true}, each with a different random initial condition. Each trained reservoir makes predictions on an ensemble of testing time series data sets, which again each have a different random initial condition. The duration and number of each of these training and testing time series data sets, along with the number of reservoir realizations tested, can be found in Table~\ref{table:testing_params}. Panel (A) in Fig.~\ref{fig:ks_stab_histogram} shows a histogram of the mean map error values of predictions generated from a reservoir computer trained using Jacobian and Tikhonov regularization using the regularization parameter values contained in Table~\ref{table:test_grid}. The histogram is multimodal, with the clearest division between cases with $\bar{\epsilon}_{map} < 10$ and $\bar{\epsilon}_{map} > 10$. Some cases with $1 < \bar{\epsilon}_{map} < 10$ stay in the general vicinity of the attractor but do not reproduce its climate, as illustrated by the $3^{\text{rd}}$ example in Fig.~\ref{fig:ks_stab_histogram}; we have chosen to classify these as unstable. Our choice for a cutoff of $\bar{\epsilon}_{map} = 1$ between cases we classify as stable or unstable is meant to ensure that those we classify as stable do produce a reasonably accurate climate, as illustrated by the first 2 examples in Fig.~\ref{fig:ks_stab_histogram}. We see that the predictions categorized as stable appear to have chaotic behavior at the end of the prediction period and have a power spectral density that matches that of the true system well, while the prediction categorized as unstable has a periodic behavior by the end of the prediction period and, as such, has a power spectral density very different from that of the true system. We additionally note that the stable prediction with a low mean map error has a power spectral density that is closer to the truth than that with a high mean map error, indicating that this metric is useful for determining how closely the prediction climate will match that of the true system.

Table~\ref{table:test_results} shows the results of reservoir predictions made using different regularization methods, ordered from worst to best performance. Figures~\ref{fig:ks_preds_unstable} and \ref{fig:ks_preds_stable} show examples of unstable and stable predictions, respectively. Each figure uses a particular reservoir computer, training time series data set, and testing time series data set drawn from our ensemble. We note that the reservoir computer and training time series data set used in each figure are different, while the testing time series data set is the same. In addition, we want to emphasize that the predictions shown in Fig.~\ref{fig:ks_preds_stable} are observed to be stable for the entire long prediction period of $192$ $t_{Lyap}$. Figure ~\ref{fig:ks_map_error} shows the map error $\epsilon_{map}(t)$ during each of the predictions shown in Figs.~\ref{fig:ks_preds_unstable} and \ref{fig:ks_preds_stable}. We observe that the map error of the unstable predictions curves increases, while the map error in the stable prediction remains small throughout the entire prediction period ($192$ $t_{Lyap}$). We note that the map error of the predictions made using noise training and Tikhonov regularization and LMNT and Tikhonov regularization are very similar for $t_{pred} < 7\;t_{Lyap}$.

We see that reservoir computers trained without regularization or using only Jacobian regularization produce predictions with a very low median valid time (less than $30\%$ of a Lyapunov time) and are always unstable. Reservoir computers trained with only Tikhonov regularization have a somewhat longer, but still poor prediction valid time, while just over half of the predictions remain stable. Furthermore, they require a high amount of regularization to reach this fraction of stability. Reservoir computers trained with Tikhonov regularization in addition to either Jacobian regularization, Noise Training, or LMNT produce predictions that are always observed to be stable and appear to remain stable for arbitrarily long times with appropriately chosen regularization amounts. Prediction valid times using the combination of Jacobian and Tikhonov regularization, however, are substantially lower than for those that are trained with the other stable regularization types. 

\subsubsection{Prediction Climate}
\begin{figure}[!ht]
    \centering
    \includegraphics[width = 0.9\textwidth]{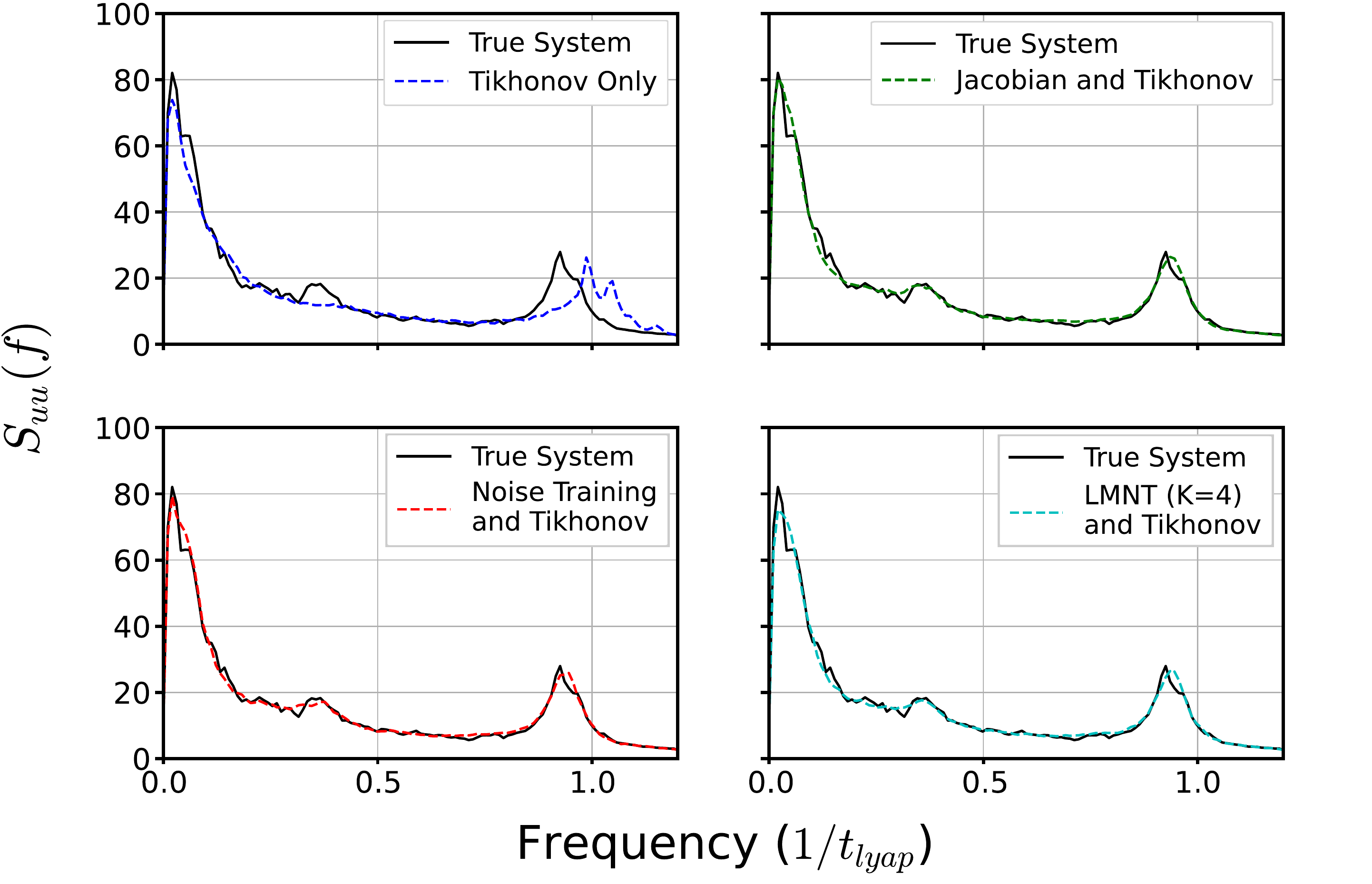}
    \caption{Average Predicted PSD Compared to True PSD. The true KS equation PSD is computed over a single trajectory of duration $T_{pred} = 1,000,000\:(12,000\;t_{Lyap})$, while the prediction PSD is computed as the average over predictions of duration $T_{pred}=16,000\:(192\;t_{Lyap})$ produced by the ensemble of reservoirs, training data sets, and testing initial conditions used to test the prediction climate. When computing the PSD with Welch's method, we used a window with $2^{13}$ samples $(98.304\: t_{lyap})$.}
    \label{fig:ks_power_spectrum}
\end{figure}

As discussed, results for some regularization methods appear to be stable for arbitrarily long times. These are thus potential candidates for climate predictions. In Fig.~\ref{fig:ks_power_spectrum}, we display the PSD computed from time series data from the true system and from predictions made using these long-time stable results. We find that predictions using reservoir computers trained with only Tikhonov Regularization (using only those orbits that remain stable) give a useful, but imperfect, replication of the PSD of the true system climate. We observe in Fig.~\ref{fig:ks_power_spectrum} that all of the other regularization methods that resulted in stable predictions are able to capture the PSD of the climate with very high accuracy.

\subsubsection{Regularization Optimization}\label{sec:hyperparam_optim}
\begin{figure}[!ht]
    \centering
    \includegraphics[width = 0.87\textwidth]{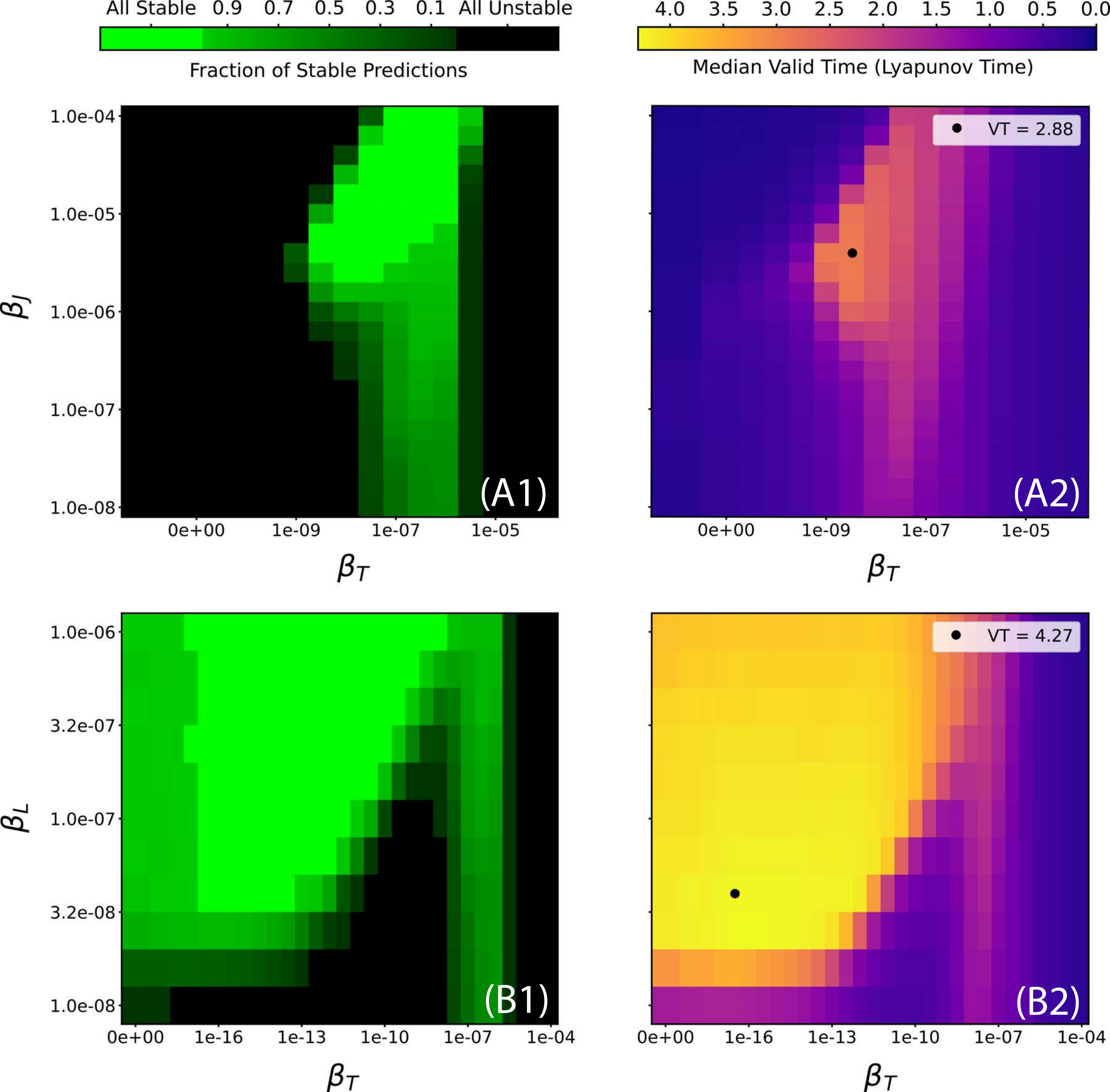}
    \caption{Fraction of Stable Predictions and Median Valid Time using Varying Regularization Parameter Values. Color plots displaying the fraction of stable predictions and median prediction valid time over a grid of different regularization parameters values are shown on the left and right, respectively. Panels (A1) and (A2) are generated from predictions made from reservoir computers trained with Jacobian and Tikhonov regularization; the regularization parameters are plotted along the vertical and horizontal axes, respectively. Panels (B1) and (B2) are generated from predictions made from reservoir computers trained with LMNT and Tikhonov regularization; again, the regularization parameters are plotted along the vertical and horizontal axes, respectively. The black dots in panels (A2) and (B2) mark the regularization parameter values chosen to both fully stabilize the predictions and produce the largest median prediction valid time, which is noted in the white box in the top right of each panel.}
    \label{fig:ks_reg_optim}
\end{figure}
In this section, we show how the optimal regularization parameter values used to produce the results shown in  Figs.~\ref{fig:ks_preds_unstable}-\ref{fig:ks_power_spectrum} and in Table~\ref{table:test_results} are obtained. Figure~\ref{fig:ks_reg_optim} displays, for two of the regularization methods, the fraction of stable predictions and the median prediction valid time for predictions made using reservoir computers trained with regularization parameter values distributed over a logarithmic grid. For each method, among all of the tested regularization parameter values, we choose those resulting in predictions that are always stable and have the highest prediction valid time. The corresponding plots generated using noise training and Tikhonov regularization look very similar to those generated using LMNT and Tikhonov regularization (panels (B1) and (B2)).

In general, we find that the chosen regularization parameters are close to the boundary between parameters that produce predictions that are always stable and those that sometimes produce unstable predictions. This result demonstrates why it is advantageous be able to efficiently tune the regularization parameters used during training; the more efficiently one can tune, the closer one can explore to this stability boundary, and thus the better one's model will be. As discussed in Sec.~\ref{sec:noise_training}, the desirability of an easily-tunable regularization makes LMNT preferable to noise training, which requires re-computation of the reservoir internal states each time the regularization parameter $\beta_N$ is changed.

We remark that in some cases, it is possible to obtain an increased median prediction valid time at the cost of a small fraction of predictions being long-term unstable. For example, if we were to decrease $\beta_T$ from the optimal value marked by the black dot in panel (A2) in Fig.~\ref{fig:ks_reg_optim}, the median prediction valid time over all of the predictions would increase; however, panel (A1) of Fig.~\ref{fig:ks_reg_optim} indicates that some of our predictions will now eventually become unstable. In addition, we see from panel (B1) that increasing the LMNT regularization parameter value from the optimal (e.g., to $10^{-6}$) improves the robustness of our prediction stability to changes in the Tikhonov regularization parameter value. Panel (B2) shows that this comes at the cost of a decreased valid prediction time; however, if one is only able to perform a coarse hyperparameter optimization, then choosing a more robust LMNT regularization parameter value may be necessary. 
These cases emphasize how different choices of regularization parameter values can be optimal for different tasks.
\section{Discussion and Conclusion}
\label{sec:conclusion}
In the absence of mitigating techniques, long-term forecasting using the machine learning approach shown in Fig.~\ref{fig:basicML} can often become unstable. We show that in the case of reservoir computing, machine learning can be trained to produce stable predictions of a paradigmatic chaotic test system, the Kuramoto-Sivashinsky equation, by training with added input noise or with our new regularization technique, LMNT. We find that reservoir computers trained using only Tikhonov regularization are only able to produce stable predictions for some reservoir realizations and training data sets. Reservoir computers trained with Jacobian and Tikhonov regularization make substantially less accurate short-term forecasts, measured by prediction valid time, than those trained with noise training or LMNT. In addition, we find that all stable predictions are able to accurately reproduce the climate of the Kuramoto-Sivashinsky equation, though reservoir computers trained with only Tikhonov regularization did not reproduce the climate as accurately as the other techniques used. In the LMNT case, the regularization matrix need not be computed again for each regularization strength value tested during hyperparameter tuning, presenting a clear advantage over noise training, which is more difficult to tune.

While training with LMNT leads to accurate and stable predictions, computing the LMNT regularization matrix can be computationally intensive due to the many matrix-matrix products involved (see Eq.~\ref{eq:lmnt}). One way to speed up this computation during training would be to approximate the LMNT regularization matrix using a smaller number of training samples than is used for the rest of the training. Another, still more approximate and heuristic, possibility would be to compute the regularization using a constant input, such as the mean computed over the training data. We discuss our results using these two methods in~\ref{app:reduced_lmnt}.

Work remains to be completed on evaluating the performance of our new regularization technique on other systems, terrestrial climate in particular. As of the writing of this article, we have implemented a version of the reduced training sample LMNT in the hybrid atmospheric model described in Ref.~\cite{arcomano_hybrid_2022}. In preliminary results, our implementation of LMNT leads to terrestrial climate predictions that are stable for over the decade-long run duration test and maintain a good climate, similar to the results achieved using input noise. The LMNT technique may also be suitable for other RNN training methods, such as LSTM; in this context, one could also compare this technique to other regularization techniques, such as LASSO or multiple feedback training, which require that the model be trained using gradient descent or similar iterative methods.
\section*{Data and Code}
The data and code that support the findings of this article are freely available at at \href{https://github.com/awikner/res-noise-stabilization}{https://github.com/awikner/res-noise-stabilization}.
\section*{Acknowledgements}
The research completed for this article was supported by DARPA contract HR00112290035. Alexander Wikner's contribution to this research was also supported in part by NSF award DGE-1632976. Joseph Harvey's contribution to this research was supported by NSF award PHY-2150399.
\appendix
\section{LMNT Performance vs. Number of Noise Steps, $K$}\label{app:lmnt_K}
\begin{center}
\begin{table}[!ht]
    \centering
    \resizebox{\columnwidth}{!}{
    \begin{tabular}{|l||l|l|l|l|l|}
        \hline
        \multicolumn{6}{|c|}{LMNT Test Results}\\
        \hline
         Noise Steps, $K$ &  \begin{tabular}{l} Regularization\\ Parameters\end{tabular} & \begin{tabular}{l}Fraction \\of Stable \\Predictions \end{tabular}& \begin{tabular}{l} Median \\Valid Time \\ ($t_{Lyap}$) \end{tabular}& \begin{tabular}{l} Median \\$\bar{\epsilon}_{map}$ \end{tabular}& \begin{tabular}{l}Median \\$\epsilon_{map}^{max}$\end{tabular}\\ \hline\hline
         1 (Jacobian)  & \begin{tabular}{l} $\beta_J = 10^{-5.4}$ \\ $\beta_T = 10^{-8.5}$  \end{tabular} & $1000/1000$ & $2.88\pm 0.02$ &\begin{tabular}{l}$(9.16\pm 0.04)$\\$\times10^{-3}$ \end{tabular}& \begin{tabular}{l}$(5.82\pm 0.06)$\\$\times10^{-2}$ \end{tabular}\\ \hline
         2  & \begin{tabular}{l} $\beta_L = 10^{-6.4}$ \\ $\beta_T = 10^{-10.5}$ \end{tabular} & $1000/1000$ & $3.92 \pm 0.04$ & \begin{tabular}{l}$(4.10\pm 0.02)$\\$\times 10^{-3}$ \end{tabular} & \begin{tabular}{l}$(2.65\pm 0.03)$\\$\times 10^{-2}$ \end{tabular}\\ \hline
         3  & \begin{tabular}{l} $\beta_L = 10^{-6.6}$ \\ $\beta_T = 10^{-12}$ \end{tabular} & $1000/1000$ & $4.04 \pm 0.04$ & \begin{tabular}{l}$(3.44\pm 0.01)$\\$\times 10^{-3}$ \end{tabular} & \begin{tabular}{l}$(2.39\pm 0.03)$\\$\times 10^{-2}$ \end{tabular}\\ \hline
         4  & \begin{tabular}{l} $\beta_L = 10^{-7.4}$ \\ $\beta_T = 10^{-16.5}$ \end{tabular} & $1000/1000$ & $4.27\pm 0.04$ & \begin{tabular}{l}$(2.75\pm 0.01)$\\$\times 10^{-3}$ \end{tabular}& \begin{tabular}{l}$(2.03\pm 0.04)$\\$\times 10^{-2}$\end{tabular}\\ \hline
         5 & \begin{tabular}{l} $\beta_L = 10^{-7.4}$ \\ $\beta_T = 10^{-16.5}$ \end{tabular} & $1000/1000$ & $4.27 \pm 0.04$ & \begin{tabular}{l}$(2.73\pm 0.02)$\\$\times 10^{-3}$ \end{tabular} & \begin{tabular}{l}$(2.02\pm 0.02)$\\$\times 10^{-2}$ \end{tabular}\\ \hline
    \end{tabular}
    }
    \caption{Reservoir computer prediction results using Tikhonov regularization and LMNT regularization, where we vary the number of noise steps used to compute the LMNT regularization. For a detailed description of how we select regularization parameter values and compute uncertainty bounds, see the Table~\ref{table:test_results} caption.}
    \label{table:lmnt_K_results}
    \end{table}
\end{center}
Computing the LMNT regularization matrix can be computationally expensive due to the many matrix-matrix products and sums involved (see Eq.~\ref{eq:lmnt}). In our reservoir computing implementation, this high computational cost is mitigated by the fact that $\vv{A}$ and all diagonal matrices $\di{\dots}$ are sparse; nevertheless, for computational efficiency one should choose the number of noise steps that the LMNT regularization approximates, $K$, to be as small as possible while still resulting in near-optimal prediction stability and prediction accuracy. Table~\ref{table:lmnt_K_results} displays the results of reservoirs trained with Tikhonov and LMNT regularization as we increase the number of noise steps approximated. We see that while we can select regularization parameter values such that all predictions are stable for all values of $K$ tested, the median valid time increases and the median $\bar{\epsilon}_{map}$ and median $\epsilon_{map}^{max}$ decreases as $K$ is increased. We find that, in this scenario, $K=4$ gives as high a median valid time as low a median $\bar{\epsilon}_{map}$ and median $\epsilon_{map}^{max}$ as higher $K$; we have therefore chosen to use $K=4$ for all other tests of the LMNT regularization discussed in this article.
\section{LMNT Performance with Reduced Training Data}
\label{app:reduced_lmnt}
While it is most natural to compute the LMNT regularization using all of the training states (similar to how we compute the Jacobian regularization), this computation becomes expensive for long training times and large $K$ values, as discussed in~\ref{app:lmnt_K}. One can substantially decrease this computational cost and potentially obtain a useful regularization matrix by either (a) computing the regularization using a size $T$ subset of the training states, obtained by sampling the states uniformly, or (b) computing the regularization using the mean input computed over the training data and the reservoir state synchronized to this mean. The regularization matrix for the technique (a) is
\begin{align}\label{eq:reduced_lmnt}
        \vv{R}_{L,T} = &\frac{1}{T}\sum_{j=0}^{T-1}\Bigg[\sum_{k=1+\text{floor}(j\tau)}^{K+\text{floor}(j\tau)}\bb{\nabla}_\vv{u}(K+\text{floor}(j\tau),k)\bb{\nabla}_\vv{u}(K+\text{floor}(j\tau),k)^\intercal\Bigg],
\end{align}
where $\tau = (T_{train}-K)/T$, $T < T_{train} - K$ is a positive integer, and $\text{floor}(\dots)$ denotes rounding down. The regularization matrix for technique (b) is
\begin{align}\label{eq:zero_lmnt}
        \vv{R}_{L,0} = {}&\sum_{k=1}^{K}\bb{\nabla}_\vv{u,0}(K,k)\bb{\nabla}_\vv{u,0}(K,k)^\intercal.
\end{align}
In Eq.~\ref{eq:zero_lmnt}, $\bb{\nabla}_\vv{u,0}(K,k)$ denotes evaluation of the right side of Eq.~\ref{eq:lmnt_jacobian} at $\vv{u}_k = \vv{0}$ (the mean of our standardized training data) and $\vv{s}_{j-1}=\vv{s}_{j-2}=\dots=\vv{s}_k=\vv{s}_{mean}$, where $\vv{s}_{mean}$ is the reservoir feature vector synchronized to the constant mean input.

\begin{center}
\begin{table}[!ht]
    \centering
    \resizebox{\columnwidth}{!}{
    \begin{tabular}{|l||l|l|l|l|l|}
        \hline
        \multicolumn{6}{|c|}{LMNT ($K=4$) Test Results}\\
        \hline
         Training Samples, $T$ &  \begin{tabular}{l} Regularization\\ Parameters\end{tabular} & \begin{tabular}{l}Fraction \\of Stable \\Predictions \end{tabular}& \begin{tabular}{l} Median \\Valid Time \\ ($t_{Lyap}$) \end{tabular}& \begin{tabular}{l} Median \\$\bar{\epsilon}_{map}$ \end{tabular}& \begin{tabular}{l}Median \\$\epsilon_{map}^{max}$\end{tabular}\\ \hline\hline
         Mean Input (Eq.~\Ref{eq:zero_lmnt}) & \begin{tabular}{l} $\beta_L = 10^{-7.4}$ \\ $\beta_T = 10^{-15.5}$ \end{tabular} & $1000/1000$ & $4.30\pm 0.04$ & \begin{tabular}{l}$(2.65\pm 0.01)$\\$\times 10^{-3}$ \end{tabular} & \begin{tabular}{l}$(2.00\pm 0.03)$\\$\times 10^{-2}$ \end{tabular} \\ \hline
         1 & \begin{tabular}{l} $\beta_L = 10^{-5.6}$ \\ $\beta_T = 10^{-10}$ \end{tabular} & $1000/1000$ & $3.74 \pm 0.07$ & \begin{tabular}{l}$(5.57\pm 0.03)$\\$\times 10^{-3}$ \end{tabular} & \begin{tabular}{l}$(3.46\pm 0.04)$\\$\times 10^{-2}$ \end{tabular}\\ \hline
         5 & \begin{tabular}{l} $\beta_L = 10^{-7.2}$ \\ $\beta_T = 10^{-16.5}$ \end{tabular} & $1000/1000$ & $4.21 \pm 0.04$ & \begin{tabular}{l}$(2.89\pm 0.02)$\\$\times 10^{-3}$ \end{tabular} & \begin{tabular}{l}$(2.16\pm 0.03)$\\$\times 10^{-2}$ \end{tabular}\\ \hline
         10 & \begin{tabular}{l} $\beta_L = 10^{-7.2}$ \\ $\beta_T = 10^{-16}$ \end{tabular} & $1000/1000$ & $4.21 \pm 0.04$ & \begin{tabular}{l}$(2.89\pm 0.02)$\\$\times 10^{-3}$ \end{tabular} & \begin{tabular}{l}$(2.12\pm 0.03)$\\$\times 10^{-2}$ \end{tabular}\\ \hline
         20& \begin{tabular}{l} $\beta_L = 10^{-7.4}$ \\ $\beta_T = 10^{-16.5}$ \end{tabular} & $1000/1000$ & $4.26 \pm 0.04$ & \begin{tabular}{l}$(2.75\pm 0.02)$\\$\times 10^{-3}$ \end{tabular} & \begin{tabular}{l}$(2.03\pm 0.03)$\\$\times 10^{-2}$ \end{tabular}\\ \hline
         100 & \begin{tabular}{l} $\beta_L = 10^{-7.4}$ \\ $\beta_T = 10^{-16.5}$ \end{tabular} & $1000/1000$ & $4.27 \pm 0.04$ & \begin{tabular}{l}$(2.73\pm 0.02)$\\$\times 10^{-3}$ \end{tabular} & \begin{tabular}{l}$(2.03\pm 0.03)$\\$\times 10^{-2}$ \end{tabular}\\ \hline
         19,996 & \begin{tabular}{l} $\beta_L = 10^{-7.4}$ \\ $\beta_T = 10^{-16.5}$ \end{tabular} & $1000/1000$ & $4.27\pm 0.04$ & \begin{tabular}{l}$(2.75\pm 0.01)$\\$\times 10^{-3}$ \end{tabular}& \begin{tabular}{l}$(2.03\pm 0.04)$\\$\times 10^{-2}$\end{tabular}\\ \hline
    \end{tabular}
    }
    \caption{Reservoir computer prediction results using LMNT regularization computed with different numbers of training samples. For a detailed description of how we select regularization parameter values and compute uncertainty bounds, see the Table~\ref{table:test_results} caption.}
    \label{table:lmnt_test_results}
    \end{table}
\end{center}
Table~\ref{table:lmnt_test_results} shows the prediction results obtained from reservoir trained with LMNT, reduced training sample LMNT, and mean-input LMNT. For our particular test system, we find that training with only $20$ training samples ($0.1\%$ of the available training data) is sufficient to obtain predictions that are always observed to be stable and have a median valid time equivalent to that when the LMNT regularization is computed with all of the training data. This effectiveness for a small number of training samples indicates that LMNT-like regularization would also be effective in models trained in batches using stochastic gradient descent or its derivatives. In addition, computing the LMNT regularization with the mean input and synchronized reservoir state performs as well as the full LMNT regularization. This result suggests that, in this case, the variability across the training data of the derivative matrices $\bb{\nabla}_\vv{u}(j,k)$ is small enough that a small number of samples, or one representative derivative matrix, is sufficient to compute a regularization matrix that performs comparably to the full LMNT.  If this property holds in other applications, the computationally simpler mean-input LMNT might yield sufficient stabilization and make the full or partial LMNT computation unnecessary.

\bibliography{bibliography_db}

\providecommand{\noopsort}[1]{}
\begin{thebibliography}{10}
\expandafter\ifx\csname url\endcsname\relax
  \def\url#1{\texttt{#1}}\fi
\expandafter\ifx\csname urlprefix\endcsname\relax\def\urlprefix{URL }\fi
\expandafter\ifx\csname href\endcsname\relax
  \def\href#1#2{#2} \def\path#1{#1}\fi

\bibitem{rasp_weatherbench_2020}
S.~Rasp, P.~D. Dueben, S.~Scher, J.~A. Weyn, S.~Mouatadid, N.~Thuerey,
  {{WeatherBench}}: {{A Benchmark Data Set}} for {{Data-Driven Weather
  Forecasting}}, Journal of Advances in Modeling Earth Systems 12~(11) (2020)
  e2020MS002203.
\newblock \href {https://doi.org/10.1029/2020MS002203}
  {\path{doi:10.1029/2020MS002203}}.

\bibitem{arcomano_machine_2020}
T.~Arcomano, I.~Szunyogh, J.~Pathak, A.~Wikner, B.~R. Hunt, E.~Ott, A {{Machine
  Learning-Based Global Atmospheric Forecast Model}}, Geophysical Research
  Letters 47~(9) (2020) e2020GL087776.
\newblock \href {https://doi.org/10.1029/2020GL087776}
  {\path{doi:10.1029/2020GL087776}}.

\bibitem{rasp_data-driven_2021}
S.~Rasp, N.~Thuerey, Data-{{Driven Medium-Range Weather Prediction With}} a
  {{Resnet Pretrained}} on {{Climate Simulations}}: {{A New Model}} for
  {{WeatherBench}}, Journal of Advances in Modeling Earth Systems 13~(2) (2021)
  e2020MS002405.
\newblock \href {https://doi.org/10.1029/2020MS002405}
  {\path{doi:10.1029/2020MS002405}}.

\bibitem{pathak_fourcastnet_2022}
J.~Pathak, S.~Subramanian, P.~Harrington, S.~Raja, A.~Chattopadhyay,
  M.~Mardani, T.~Kurth, D.~Hall, Z.~Li, K.~Azizzadenesheli, P.~Hassanzadeh,
  K.~Kashinath, A.~Anandkumar, {{FourCastNet}}: {{A Global Data-driven
  High-resolution Weather Model}} using {{Adaptive Fourier Neural Operators}}
  (Feb. 2022).
\newblock \href {http://arxiv.org/abs/2202.11214} {\path{arXiv:2202.11214}},
  \href {https://doi.org/10.48550/arXiv.2202.11214}
  {\path{doi:10.48550/arXiv.2202.11214}}.

\bibitem{pathak_model-free_2018}
J.~Pathak, B.~Hunt, M.~Girvan, Z.~Lu, E.~Ott, Model-{{Free Prediction}} of
  {{Large Spatiotemporally Chaotic Systems}} from {{Data}}: {{A Reservoir
  Computing Approach}}, Phys. Rev. Lett. 120~(2) (2018) 024102.
\newblock \href {https://doi.org/10.1103/PhysRevLett.120.024102}
  {\path{doi:10.1103/PhysRevLett.120.024102}}.

\bibitem{vlachas_data-driven_2018}
P.~R. Vlachas, W.~Byeon, Z.~Y. Wan, T.~P. Sapsis, P.~Koumoutsakos, Data-driven
  forecasting of high-dimensional chaotic systems with long short-term memory
  networks, Proceedings of the Royal Society A: Mathematical, Physical and
  Engineering Sciences 474~(2213) (2018) 20170844.
\newblock \href {https://doi.org/10.1098/rspa.2017.0844}
  {\path{doi:10.1098/rspa.2017.0844}}.

\bibitem{vlachas_backpropagation_2020}
P.~R. Vlachas, J.~Pathak, B.~R. Hunt, T.~P. Sapsis, M.~Girvan, E.~Ott,
  P.~Koumoutsakos, Backpropagation algorithms and {{Reservoir Computing}} in
  {{Recurrent Neural Networks}} for the forecasting of complex spatiotemporal
  dynamics, Neural Networks 126 (2020) 191--217.
\newblock \href {https://doi.org/10.1016/j.neunet.2020.02.016}
  {\path{doi:10.1016/j.neunet.2020.02.016}}.

\bibitem{balakrishnan_deep_2020}
K.~Balakrishnan, D.~Upadhyay, Deep {{Adversarial Koopman Model}} for
  {{Reaction-Diffusion}} systems, arXiv:2006.05547 [cs, eess] (Jun. 2020).
\newblock \href {http://arxiv.org/abs/2006.05547} {\path{arXiv:2006.05547}}.

\bibitem{rasp_deep_2018}
S.~Rasp, M.~S. Pritchard, P.~Gentine, Deep learning to represent subgrid
  processes in climate models, Proceedings of the National Academy of Sciences
  115~(39) (2018) 9684--9689.
\newblock \href {https://doi.org/10.1073/pnas.1810286115}
  {\path{doi:10.1073/pnas.1810286115}}.

\bibitem{gentine_could_2018}
P.~Gentine, M.~Pritchard, S.~Rasp, G.~Reinaudi, G.~Yacalis, Could {{Machine
  Learning Break}} the {{Convection Parameterization Deadlock}}?, Geophysical
  Research Letters 45~(11) (2018) 5742--5751.
\newblock \href {https://doi.org/10.1029/2018GL078202}
  {\path{doi:10.1029/2018GL078202}}.

\bibitem{li_fourier_2021}
Z.~Li, N.~Kovachki, K.~Azizzadenesheli, B.~Liu, K.~Bhattacharya, A.~Stuart,
  A.~Anandkumar, Fourier {{Neural Operator}} for {{Parametric Partial
  Differential Equations}} (May 2021).
\newblock \href {http://arxiv.org/abs/2010.08895} {\path{arXiv:2010.08895}},
  \href {https://doi.org/10.48550/arXiv.2010.08895}
  {\path{doi:10.48550/arXiv.2010.08895}}.

\bibitem{auslander_attractors_1964}
J.~Auslander, N.~P. Bhatia, P.~Seibert, Attractors in dynamical systems, Tech.
  Rep. NASA-CR-59858, {NASA} (Jan. 1964).

\bibitem{lu_attractor_2018}
Z.~Lu, B.~R. Hunt, E.~Ott, Attractor reconstruction by machine learning, Chaos
  28~(6) (2018) 061104.
\newblock \href {https://doi.org/10.1063/1.5039508}
  {\path{doi:10.1063/1.5039508}}.

\bibitem{pathak_using_2017}
J.~Pathak, Z.~Lu, B.~R. Hunt, M.~Girvan, E.~Ott, Using machine learning to
  replicate chaotic attractors and calculate {{Lyapunov}} exponents from data,
  Chaos 27~(12) (2017) 121102.
\newblock \href {https://doi.org/10.1063/1.5010300}
  {\path{doi:10.1063/1.5010300}}.

\bibitem{gauthier_learning_2022}
D.~J. Gauthier, I.~Fischer, A.~R{\"o}hm, Learning unseen coexisting attractors,
  Chaos 32~(11) (2022) 113107.
\newblock \href {https://doi.org/10.1063/5.0116784}
  {\path{doi:10.1063/5.0116784}}.

\bibitem{rohm_model-free_2021}
A.~R{\"o}hm, D.~J. Gauthier, I.~Fischer, Model-free inference of unseen
  attractors: {{Reconstructing}} phase space features from a single noisy
  trajectory using reservoir computing, Chaos 31~(10) (2021) 103127.
\newblock \href {https://doi.org/10.1063/5.0065813}
  {\path{doi:10.1063/5.0065813}}.

\bibitem{jaeger_echo_2001}
H.~Jaeger, The ``echo state'' approach to analysing and training recurrent
  neural networks-with an erratum note, Bonn, Germany: German National Research
  Center for Information Technology GMD Technical Report 148 (Jan. 2001).

\bibitem{lukosevicius_reservoir_2009-1}
M.~Luko{\v s}evi{\v c}ius, H.~Jaeger, Reservoir computing approaches to
  recurrent neural network training, Computer Science Review 3~(3) (2009)
  127--149.
\newblock \href {https://doi.org/10.1016/j.cosrev.2009.03.005}
  {\path{doi:10.1016/j.cosrev.2009.03.005}}.

\bibitem{tikhonov_solutions_1977}
A.~Tikhonov, V.~Arsenin, Solutions of Ill-Posed Problems, Scripta Series in
  Mathematics, {Winston}, 1977.

\bibitem{tibshirani_regression_1996}
R.~Tibshirani, Regression {{Shrinkage}} and {{Selection Via}} the {{Lasso}},
  Journal of the Royal Statistical Society: Series B (Methodological) 58~(1)
  (1996) 267--288.
\newblock \href {https://doi.org/10.1111/j.2517-6161.1996.tb02080.x}
  {\path{doi:10.1111/j.2517-6161.1996.tb02080.x}}.

\bibitem{hoffman_robust_2019}
J.~Hoffman, D.~A. Roberts, S.~Yaida, Robust {{Learning}} with {{Jacobian
  Regularization}} (Aug. 2019).
\newblock \href {http://arxiv.org/abs/1908.02729} {\path{arXiv:1908.02729}},
  \href {https://doi.org/10.48550/arXiv.1908.02729}
  {\path{doi:10.48550/arXiv.1908.02729}}.

\bibitem{srivastava_dropout_2014}
N.~Srivastava, G.~Hinton, A.~Krizhevsky, I.~Sutskever, R.~Salakhutdinov,
  Dropout: {{A Simple Way}} to {{Prevent Neural Networks}} from
  {{Overfitting}}, The journal of machine learning research 15~(1) (2014)
  1929--1958.

\bibitem{kolen_field_2001}
J.~F. Kolen, S.~C. Kremer, A {{Field Guide}} to {{Dynamical Recurrent
  Networks}}, {John Wiley \& Sons}, 2001.

\bibitem{lamb_professor_2016}
A.~M. Lamb, A.~Goyal, Y.~Zhang, S.~Zhang, A.~C. Courville, Y.~Bengio, Professor
  {{Forcing}}: {{A New Algorithm}} for {{Training Recurrent Networks}}, in:
  Advances in {{Neural Information Processing Systems}}, Vol.~29, {Curran
  Associates, Inc.}, 2016.

\bibitem{sussillo_generating_2009}
D.~Sussillo, L.~F. Abbott, Generating {{Coherent Patterns}} of {{Activity}}
  from {{Chaotic Neural Networks}}, Neuron 63~(4) (2009) 544--557.
\newblock \href {https://doi.org/10.1016/j.neuron.2009.07.018}
  {\path{doi:10.1016/j.neuron.2009.07.018}}.

\bibitem{lim_noisy_2021}
S.~H. Lim, N.~B. Erichson, L.~Hodgkinson, M.~W. Mahoney, Noisy {{Recurrent
  Neural Networks}}, in: Advances in {{Neural Information Processing Systems}},
  Vol.~34, {Curran Associates, Inc.}, 2021, pp. 5124--5137.

\bibitem{wikner_combining_2020}
A.~Wikner, J.~Pathak, B.~Hunt, M.~Girvan, T.~Arcomano, I.~Szunyogh,
  A.~Pomerance, E.~Ott, Combining machine learning with knowledge-based
  modeling for scalable forecasting and subgrid-scale closure of large,
  complex, spatiotemporal systems, Chaos 30~(5) (2020) 053111.
\newblock \href {https://doi.org/10.1063/5.0005541}
  {\path{doi:10.1063/5.0005541}}.

\bibitem{arcomano_hybrid_2022}
T.~Arcomano, I.~Szunyogh, A.~Wikner, J.~Pathak, B.~R. Hunt, E.~Ott, A {{Hybrid
  Approach}} to {{Atmospheric Modeling That Combines Machine Learning With}} a
  {{Physics-Based Numerical Model}}, Journal of Advances in Modeling Earth
  Systems 14~(3) (2022) e2021MS002712.
\newblock \href {https://doi.org/10.1029/2021MS002712}
  {\path{doi:10.1029/2021MS002712}}.

\bibitem{hochreiter_long_1997}
S.~Hochreiter, J.~Schmidhuber, Long {{Short-Term Memory}}, Neural Computation
  9~(8) (1997) 1735--1780.
\newblock \href {https://doi.org/10.1162/neco.1997.9.8.1735}
  {\path{doi:10.1162/neco.1997.9.8.1735}}.

\bibitem{cho_properties_2014}
K.~Cho, B.~{\noopsort{merrienboer}}{van Merrienboer}, D.~Bahdanau, Y.~Bengio,
  On the {{Properties}} of {{Neural Machine Translation}}: {{Encoder-Decoder
  Approaches}} (Oct. 2014).
\newblock \href {http://arxiv.org/abs/1409.1259} {\path{arXiv:1409.1259}},
  \href {https://doi.org/10.48550/arXiv.1409.1259}
  {\path{doi:10.48550/arXiv.1409.1259}}.

\bibitem{pecora_synchronization_2015}
L.~M. Pecora, T.~L. Carroll, Synchronization of chaotic systems, Chaos 25~(9)
  (2015) 097611.
\newblock \href {https://doi.org/10.1063/1.4917383}
  {\path{doi:10.1063/1.4917383}}.

\bibitem{daubechies_iterative_2004}
I.~Daubechies, M.~Defrise, C.~De~Mol, An iterative thresholding algorithm for
  linear inverse problems with a sparsity constraint, Communications on Pure
  and Applied Mathematics 57~(11) (2004) 1413--1457.
\newblock \href {https://doi.org/10.1002/cpa.20042}
  {\path{doi:10.1002/cpa.20042}}.

\bibitem{laug}
E.~Anderson, Z.~Bai, C.~Bischof, S.~Blackford, J.~Demmel, J.~Dongarra,
  J.~Du~Croz, A.~Greenbaum, S.~Hammarling, A.~McKenney, D.~Sorensen, {{LAPACK}}
  Users' Guide, 3rd Edition, {Society for Industrial and Applied Mathematics},
  {Philadelphia, PA}, 1999.

\bibitem{harris2020array}
C.~R. Harris, K.~J. Millman, S.~J. {\noopsort{walt}}{van der Walt}, R.~Gommers,
  P.~Virtanen, D.~Cournapeau, E.~Wieser, J.~Taylor, S.~Berg, N.~J. Smith,
  R.~Kern, M.~Picus, S.~Hoyer, M.~H. {\noopsort{kerkwijk}}{van Kerkwijk},
  M.~Brett, A.~Haldane, J.~F. {\noopsort{r{\'i}o}}{del R{\'i}o}, M.~Wiebe,
  P.~Peterson, P.~{G{\'e}rard-Marchant}, K.~Sheppard, T.~Reddy, W.~Weckesser,
  H.~Abbasi, C.~Gohlke, T.~E. Oliphant, Array programming with {{NumPy}},
  Nature 585~(7825) (2020) 357--362.
\newblock \href {https://doi.org/10.1038/s41586-020-2649-2}
  {\path{doi:10.1038/s41586-020-2649-2}}.

\bibitem{james_introduction_2014}
G.~James, D.~Witten, T.~Hastie, R.~Tibshirani, An {{Introduction}} to
  {{Statistical Learning}}: With {{Applications}} in {{R}}, {Springer New
  York}, 2014.

\bibitem{welch_use_1967}
P.~Welch, The use of fast {{Fourier}} transform for the estimation of power
  spectra: {{A}} method based on time averaging over short, modified
  periodograms, IEEE Transactions on Audio and Electroacoustics 15~(2) (1967)
  70--73.
\newblock \href {https://doi.org/10.1109/TAU.1967.1161901}
  {\path{doi:10.1109/TAU.1967.1161901}}.

\bibitem{kuramoto_diffusion-induced_1978}
Y.~Kuramoto, Diffusion-{{Induced Chaos}} in {{Reaction Systems}}, Progress of
  Theoretical Physics Supplement 64 (1978) 346--367.
\newblock \href {https://doi.org/10.1143/PTPS.64.346}
  {\path{doi:10.1143/PTPS.64.346}}.

\bibitem{sivashinsky_nonlinear_1977}
G.~I. Sivashinsky, Nonlinear analysis of hydrodynamic instability in laminar
  flames\textemdash{{I}}. {{Derivation}} of basic equations, Acta Astronautica
  4~(11) (1977) 1177--1206.
\newblock \href {https://doi.org/10.1016/0094-5765(77)90096-0}
  {\path{doi:10.1016/0094-5765(77)90096-0}}.

\bibitem{cox_exponential_2002-1}
S.~M. Cox, P.~C. Matthews, Exponential {{Time Differencing}} for {{Stiff
  Systems}}, Journal of Computational Physics 176~(2) (2002) 430--455.
\newblock \href {https://doi.org/10.1006/jcph.2002.6995}
  {\path{doi:10.1006/jcph.2002.6995}}.

\bibitem{kassam_fourth-order_2005}
A.-K. Kassam, L.~N. Trefethen, Fourth-{{Order Time-Stepping}} for {{Stiff
  PDEs}}, SIAM J. Sci. Comput. 26~(4) (2005) 1214--1233.
\newblock \href {https://doi.org/10.1137/S1064827502410633}
  {\path{doi:10.1137/S1064827502410633}}.

\bibitem{bennetin_lyapunov_1980}
G.~Bennetin, L.~Galgani, A.~Giorgilli, J.-M. Strelcyn, Lyapunov characteristic
  exponents for smooth dynamical systems and for {{Hamiltonian}} systems: {{A}}
  method for computing all of them, Meccanica 15~(9) (1980) 27.

\bibitem{lorenz_deterministic_1963}
E.~N. Lorenz, Deterministic {{Nonperiodic Flow}}, Journal of Atmospheric
  Sciences 20~(2) (1963) 130--141.
\newblock \href {https://doi.org/10.1175/1520-0469(1963)020<0130:DNF>2.0.CO;2}
  {\path{doi:10.1175/1520-0469(1963)020<0130:DNF>2.0.CO;2}}.

\bibitem{conover_practical_1999}
W.~J. Conover, Practical {{Nonparametric Statistics}}, 3rd Edition, {Wiley},
  {New York, NY, USA}, 1999.

\end{thebibliography}

\end{document}